\documentclass[sigconf]{acmart}
\usepackage{booktabs}
\usepackage{amsmath}
\usepackage{algorithm}
\usepackage{algpseudocode}
\usepackage{multirow}
\usepackage{mathtools}
\usepackage{subfigure}
\usepackage{graphicx}
\usepackage{balance}
\usepackage{color}
\usepackage{enumitem}
\usepackage{balance}

\renewcommand{\vec}[1]{\ensuremath{\mathbf{#1}}}
\newcommand{\stitle}[1]{\vspace{2mm} \noindent {\bf #1}}
\newcommand{\eg}{{\it e.g.}}

\newcommand{\ie}{{\it i.e.}}

\newcommand{\wrt}{w.r.t. }
\newcommand{\tr}{\ensuremath{^\text{tr}}}
\newcommand{\te}{\ensuremath{^\text{te}}}

\newcommand{\bG}{\ensuremath{\mathcal{G}}}
\newcommand{\bN}{\ensuremath{\mathcal{N}}}

\newcommand{\bM}{\ensuremath{\mathcal{M}}}

\newcommand{\ps}{\phantom{0}}

\AtBeginDocument{%
  \providecommand\BibTeX{{%
    \normalfont B\kern-0.5em{\scshape i\kern-0.25em b}\kern-0.8em\TeX}}}

\copyrightyear{2021} 
\acmYear{2021} 
\setcopyright{acmlicensed}\acmConference[SIGIR '21]{Proceedings of the 44th International ACM SIGIR Conference on Research and Development in Information Retrieval}{July 11--15, 2021}{Virtual Event, Canada}
\acmBooktitle{Proceedings of the 44th International ACM SIGIR Conference on Research and Development in Information Retrieval (SIGIR '21), July 11--15, 2021, Virtual Event, Canada}
\acmPrice{15.00}
\acmDOI{10.1145/3404835.3462915}
\acmISBN{978-1-4503-8037-9/21/07}

\begin{document}
\title{Meta-Inductive Node Classification across Graphs}



\author{Zhihao Wen}
\affiliation{%
  \institution{Singapore Management University}
   \country{Singapore}
   }
\email{zhwen.2019@smu.edu.sg}

\author{Yuan Fang}
\affiliation{%
  \institution{Singapore Management University}
  \country{Singapore}
  }
\email{yfang@smu.edu.sg}

\author{Zemin Liu}
\affiliation{%
  \institution{Singapore Management University}
  \country{Singapore}
  }
\email{zmliu@smu.edu.sg}


\begin{abstract}
Semi-supervised node classification on graphs is an important research problem, with many real-world applications in information retrieval such as content classification on a social network and query intent classification on an e-commerce query graph. 
While traditional approaches are largely transductive, recent graph neural networks (GNNs) integrate node features with network structures, thus enabling inductive node classification models that can be applied to new nodes or even new graphs in the same feature space. 
However, inter-graph differences still exist across graphs within the same domain. Thus, training just one global model (\eg, a state-of-the-art GNN) to handle all new graphs, whilst ignoring the inter-graph differences, can lead to suboptimal performance.

In this paper, we study the problem of inductive node classification across graphs. Unlike existing one-model-fits-all approaches, we propose a novel meta-inductive framework called MI-GNN to customize the inductive model to each graph under a meta-learning paradigm. That is, MI-GNN does not directly learn an inductive model; it learns the \emph{general knowledge} of how to train a model for semi-supervised node classification on new graphs. To cope with the differences across graphs, MI-GNN employs a \emph{dual adaptation mechanism} at both the graph and task levels. More specifically, we learn a \emph{graph prior} to adapt for the graph-level differences, and a \emph{task prior} to adapt for the task-level differences conditioned on a graph. 
Extensive experiments on five real-world graph collections demonstrate the effectiveness of our proposed model.

\end{abstract}

\ccsdesc[500]{Computing methodologies~Learning latent representations}
\ccsdesc[500]{Information systems~Data mining}
\keywords{Graph neural networks, semi-supervised node classification, inductive graph model, meta-learning}



\maketitle

\makeatletter
\def\algbackskip{\hskip-\ALG@thistlm}
\makeatother

\section{Introduction}

 Graph-structured data widely exist in diverse real-world scenarios, such as social networks, e-commerce graphs, citation graphs, and biological networks. Analysis of these graphs can uncover valuable insights about their respective application domain. In particular, semi-supervised node classification on graphs \cite{blum_learning_2001} is an important task in information retrieval. For instance, on a content-sharing social network such as Flickr, content classification enables topical filtering and tag-based retrieval for multimedia items \cite{seah2018killing}; 
on a query graph for e-commerce, query intent classification  enhances the ranking of results by focusing on the intended product category \cite{DBLP:conf/sigir/FangHC12}.
Such scenarios are semi-supervised, as only some of the nodes on the graph are \emph{labeled} with a category, whilst the remaining nodes are \emph{unlabeled}. The labeled nodes and the intrinsic structures between both labeled and unlabeled nodes (\ie, the graph) can be used for training a model to classify the unlabeled nodes.

\begin{figure*}[t]
  \centering
    \includegraphics[scale=0.65]{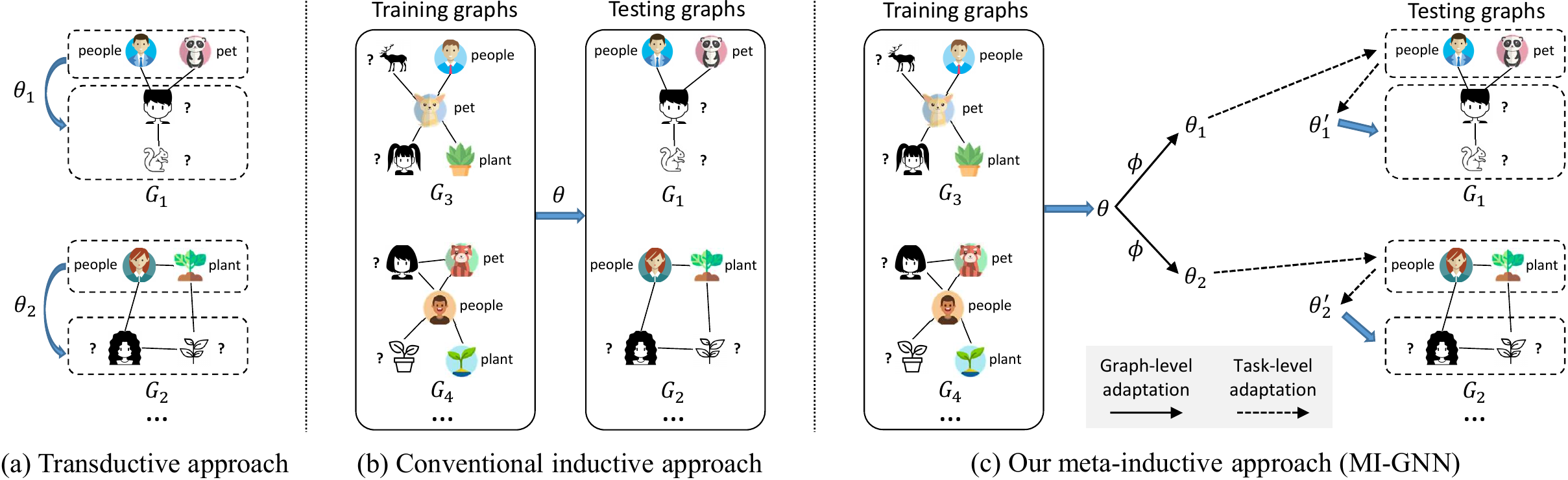}
    \vspace{-2mm}
  \caption{Illustrative comparison of transductive, inductive and our meta-inductive approaches for semi-supervised node classification on subgraphs of an image-sharing network. (Colored images: labeled nodes; black \& white images: unlabeled nodes.)}
  \label{fig:motivation}
\end{figure*}

 Unfortunately, traditional manifold-based semi-supervised approaches on graphs \cite{blum_learning_2001,zhu_semi-supervised_2003,zhou_learning_2003,wu2012learning,DBLP:conf/icml/FangCL14} mostly assume a transductive setting. That is, the learned model only works on existing nodes in the same graph, and cannot be applied to new nodes added to the existing graph or entirely new graphs even if they are from the same domain. As  Figure~\ref{fig:motivation}(a) shows, a transductive approach directly trains a model $\theta_i$ on the labeled nodes of each graph $G_i$, and apply the model to classify the unlabeled nodes in the same graph $G_i$.
While some simple inductive extensions exist through nearest neighbors or kernel regression \cite{he2007graph}, they can only deal with new nodes in a limited manner by processing the local changes, and often cannot generalize to handling new graph structures. The ability to handle new graphs is important, as we often need to deal with a series of ego-networks or subgraphs  \cite{zhang2013social,li2014user,dougnon2016inferring} when the full graph is too large to process or impossible to obtain. Thus, it becomes imperative to equip semi-supervised node classification with the inductive capability of generalizing across graphs. 

\stitle{Problem setting.}
In this paper, we study the problem of \emph{inductive semi-supervised node classification across graphs}. Consider a set of training (existing) graphs and a  set of testing (new) graphs. In a training graph, some or all of the nodes are labeled with a category; in a testing graph only some of the nodes are labeled and the rest are unlabeled. The nodes in all graphs reside in the same feature space and share a common set of categories.
Our goal is to learn an inductive model from the training graphs, which can be applied to  the testing graphs to classify their unlabeled nodes. 

\stitle{Prior work.}
State-of-the-art node classification approaches hinge on graph representation learning, which projects nodes to a latent, low-dimensional vector space. There exist two main factions: network embedding \cite{cai2018comprehensive}  and graph neural networks (GNNs) \cite{wu2020comprehensive}. 

On one hand, network embedding methods directly parameterize node embedding vectors and constrain them with various local structures, such as random walks in DeepWalk \cite{perozzi2014deepwalk} and node2vec \cite{grover2016node2vec}, and first- and second-order proximity in LINE \cite{tang2015line}.
Due to the direct parameterization, network embedding has limited inductive capability like the traditional manifold approaches. For instance, the online version of DeepWalk handles new nodes by incrementally processing the local random walks around them.  

On the other hand, GNNs integrate node features and structures into representation learning. They typically follow a message passing framework, in which each node receives, maps and aggregates messages (\ie, features or embeddings) from its neighboring nodes in multiple layers to generate its own embedding vector. The implication is that GNNs are parameterized by a weight matrix in each layer to map the messages from the neighboring nodes, instead of directly learning the node embedding vectors. In particular, the weight matrices give rise to the inherent inductive power of GNNs, which can be applied to similarly map and aggregate messages in a new graph given the same feature space.
As  Figure~\ref{fig:motivation}(b) shows, we can train a GNN model $\theta$ on a collection of training graphs $\{G_3,G_4\}$, which are image co-occurrence subgraphs of an image-sharing network like Flickr \cite{Zeng2020GraphSAINT:}. Specifically, in every subgraph, each node represents an image, and an edge can be formed between two images if they have certain common properties (\eg, submitted to the same gallery or taken by friends).
The learned model $\theta$ can be deployed to predict the unlabeled nodes on new testing graphs $\{G_1,G_2\}$, which are different subgraphs from the same image-sharing network. In particular, nodes in all subgraphs share the same feature space and belong to a common set of categories.

\stitle{Challenges and present work.}
While most GNNs can be inductive, ultimately they only train a single inductive model to apply on all new graphs. These one-model-fits-all approaches turn out to suffer from a major drawback, as they neglect inter-graph differences that can be crucial to new graphs.
Even graphs in the same domain often exhibit a myriad of differences. For instance, 
social ego-networks for different kind of ego-users (\eg, businesses, celebrities and regular users) show dissimilar structural patterns;
image co-occurrence subgraphs in different galleries have varying distributions of node features and categories. 
To cope with such inter-graph differences, it remains challenging to formulate an inductive approach that not only becomes aware of but also customizes to the differences across graphs. To be more specific, there are two open questions to address. 

First, \emph{how do we dynamically adjust the inductive model?} A na\"ive approach is to perform an additional fine-tuning step on the labeled nodes of the new graph. However, such a fine-tuning on new graphs is decoupled from the training step, which does not learn to deal with inter-graph differences. Thus, the two-step approach cannot adequately customize to different graphs. Instead, the training process must be made aware of inter-graph differences and further adapt to the differences across training graphs. In this paper, we resort to the \emph{meta-learning} paradigm \cite{santoro2016meta,vinyals2016matching}, in which we do not directly train an inductive model. Instead, we learn a form of general knowledge that can be quickly utilized to produce a customized inductive model for each new graph. In other words, the general knowledge encodes \emph{how to train} a model for new graphs. While meta-learning has been successfully adopted in various kinds of data including images \cite{liu2019learning}, texts \cite{hu2019few} and graphs \cite{zhou2019meta}, these approaches mainly address the few-shot learning problem, whereas our work is the first to leverage meta-learning for inductive semi-supervised node classification on graphs.

Second, more concretely, \emph{what form of general knowledge can empower semi-supervised node classification on a new graph?}
On one hand, every semi-supervised node classification task is different, which arises from different nodes and labels across tasks. On the other hand, every graph is different, providing a different context to the tasks on different graphs. 
Thus, the general knowledge should encode how to deal with both task- and graph-level differences. As Figure~\ref{fig:motivation}(c) illustrates, for task-level differences, we learn a \emph{task prior} $\theta$ that can be eventually adapted to the semi-supervised node classification task in a new graph; for graph-level differences, we learn a \emph{graph prior} $\phi$ that can first transform $\theta$ into $\theta_i$ conditioned on each graph $G_i$, before further adapting $\theta_i$ to $\theta_i'$ for the classification task on $G_i$.  
In other words, our general knowledge consists of the task prior and graph prior, amounting to a \emph{dual adaptation mechanism} on both tasks and graphs. Intuitively, the graph-level adaptation exploits the intrinsic relationship between graphs, whereas the task-level adaptation exploits the graph-conditioned relationship between tasks. This is a significant departure from existing task-based meta-learning approaches such as protonets \cite{snell2017prototypical} and MAML \cite{finn2017model}, which assumes that tasks are i.i.d.~sampled from a task distribution. In contrast, in our setting tasks are non-i.i.d. as they are sampled from and thus conditioned on different graphs.

\stitle{Contributions.}
Given the above challenges and insights for inductive semi-supervised node classification across graphs, we propose a novel Meta-Inductive framework for Graph Neural Networks (MI-GNN). To summarize, we make the following contributions.
(1) This is the first attempt to leverage the meta-learning paradigm for inductive semi-supervised node classification on graphs, which learns to train an inductive model for new graphs.
(2) We propose a novel framework MI-GNN, which employs a dual-adaptation mechanism to learn the general knowledge of training an inductive model at both the task and graph levels.
(3) We conduct extensive experiments on five real-world datasets, and demonstrate the superior inductive ability of the proposed MI-GNN.

\section{Related Work}
We investigate related work in three groups: graph neural networks, inductive graph representation learning and meta-learning.

\stitle{Graph neural networks.} 
A surge of attention has been attracted to graph neural networks (GNNs) \cite{wu2020comprehensive}. Based on they key operation of neighborhood aggregation in a message passing framework, they exploit the underlying graph structure and node features simultaneously. 
In particular, different message aggregation functions materialize different GNNs, \eg, GCN \cite{kipf2016semi} employs an aggregation roughly equivalent to mean pooling, and GAT \cite{velivckovic2017graph} employs self-attention to aggregate neighbors in a weighted manner. 
Recent works often exploit more structural information on graphs, such as graph isomorphism \cite{xu2018powerful} and node positions \cite{you2019position}.

\stitle{Inductive graph representation learning.}
Recent inductive learning on graphs are mainly based on network embedding and GNNs. For the former, some extend classical embedding approaches (\eg, skip-gram) to handle new nodes on dynamic graphs \cite{du2018dynamic,nguyen2018continuous,zuo2018embedding}, by exploiting structural information from the graph such as co-occurrence between nodes; others employ graph auto-encoders \cite{goyal2018dyngem,goyal2020dyngraph2vec} for dynamic graphs, by  
mining and reconstructing the graph structures. In general, this category of approaches only handle new nodes on the same graph, lacking the ability to extend to entirely new graphs.
In the latter category, most GNNs are inherently inductive, and can be applied to new graphs in the same feature space after training \cite{hamilton2017inductive,Xu2020Inductive}. 
In this paper, we follow the line of inductive learning based on GNNs.
More recently, a few pre-training approaches have also been devised for GNNs \cite{velickovic2019deep,hu2019strategies,lu2021learning}. While they also learn some form of transferable knowledge on the training graphs, they are  trained in a self-supervised manner, and the main objective is to learn universally good initializations for different downstream tasks. Thus, they address a problem setting that is different from ours.

\stitle{Meta-learning.} Also known as ``learning to learn,'' meta-learning \cite{snell2017prototypical,finn2017model} aims to simulate the learning strategy of humans, by learning some general knowledge across a set of learning tasks and adapting the knowledge to novel tasks. 
Generally, some approaches resort to protonets \cite{snell2017prototypical} which aim to learn a metric space for the class prototypes, and others  
apply optimization-based techniques such as model-agnostic meta-learning (MAML) \cite{finn2017model}. 
To further enhance task adaptation, the transferable knowledge can be tailored to different clusters of tasks, forming a hierarchical structure of adaptation \cite{yao2019hierarchically};  feature-specific memories can also guide the adapted model with a further bias \cite{dong2020mamo};   domain-knowledge graphs can also be leveraged to provide task-specific customization \cite{suo2020tadanet}.
Another subtype of meta-learning called hypernetwork \cite{ha2016hypernetworks,perez2018film} uses a secondary neural network to generate the weights for the target neural network, \ie, it learns to generate different weights conditioned on different input, instead of freezing the weights for all input after training in traditional neural networks.
More recently, meta-learning has also been adopted on graphs for few-shot learning, such as Meta-GNN \cite{zhou2019meta}, GFL \cite{yao2019graph}, GPN \cite{ding2020graph}, RALE \cite{liu2021relative} and meta-tail2vec \cite{liu2020towards}, which is distinct from inductive semi-supervised node classification as further elaborated in Section~\ref{sec:prelim:problem}. 
Hypernetwork-based approaches have also emerged, such as LGNN \cite{liu2021nodewise} that adapts GNN weights to different local contexts, and GNN-FiLM \cite{brockschmidt2019gnn} 
that adapts to different relations in a relational graph.

\section{PRELIMINARIES}

In this section, we first formalize the problem of inductive semi-supervised node classification across multiple graphs. We then present a  background on graph neural networks as the foundation of our approach.
Major notations are summarized in Table~\ref{table.notation}.

\begin{table}[t]
\centering
\small
\caption{List of major notations.} 
\vspace{-2mm}
\label{table.notation}
\begin{tabular}{@{}c|l@{}}
\toprule
 \textbf{Notation} & \textbf{Description} \\\midrule
$G, V, E, \vec{X}$ & a graph, its node and edge set, and node feature matrix  \\
$C$ & the set of node categories \\
$\ell$ & the label mapping function $V\to C$ \\
$\bN_v$ & the set of neighbors of node $v$\\
$\bG,\bG^\text{tr}, \bG^\text{te}$ & the set of all graphs, training graphs and testing graphs\\  
$G_i,S_i, Q_i$ & a graph $G_i$ with support set $S_i$ and query set $Q_i$\\  
$\theta,\phi$ & general knowledge: task prior $\theta$, graph prior $\phi$ \\
$\gamma_i,\beta_i$ & scaling and shifting vectors for graph $G_i$ \\
$\theta_i$ & graph $G_i$-conditioned task prior \\
$\theta'_i$ & graph $G_i$-conditioned and task adapted model \\\bottomrule
\end{tabular}
\end{table}

\begin{figure*}
  \centering
  \includegraphics[scale=0.65]{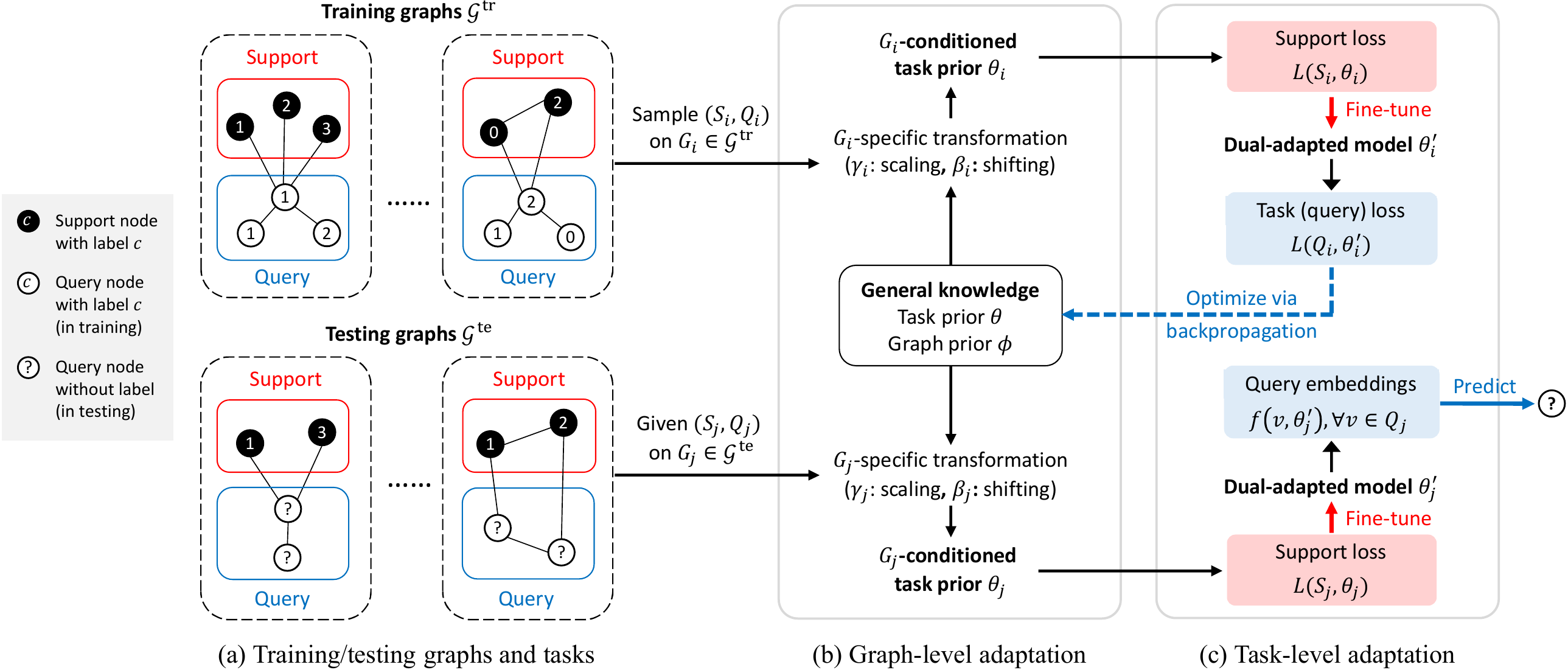}
  \vspace{-2mm}
  \caption{Overall framework of MI-GNN, illustrating the pipeline on a training graph $G_i$ and a testing graph $G_j$.}
  \label{fig:framework}
\end{figure*}

\subsection{Problem formulation}\label{sec:prelim:problem}

\stitle{Graphs.}
Our setting assumes a set of graphs $\bG$ from the same domain. A graph $G\in \bG$ is a quintuple $G=(V, E, \vec{X}, C, \ell)$ 
where (1) $V$ denotes the set of nodes; (2) $E$ denotes the set of edges between nodes; (3) $\vec{X} \in \mathbb{R}^{|V| \times d}$ is the feature matrix such that $\vec{x}_v$ is the $d$-dimensional feature vector of node $v$; (4) $C$ is the set of node categories; (5) $\ell$ is a label function that maps each node to its category, \ie, $\ell : V \to C$.
Note that the node feature space and category set are the same across all graphs.

\stitle{Inductive semi-supervised node classification.}
The graph set $\bG$ comprises two disjoint subsets, namely, training graphs $\bG^\text{tr}$ and testing graphs $\bG^{\text{te}}$. In a training graph, some or all of the nodes are labeled, \ie, the label mapping $\ell$ is known for these nodes. In contrast, in a testing graph, only some of the nodes are \emph{labeled} and the remaining nodes are \emph{unlabeled}, \ie, their label mapping is unknown.
Subsequently, given a set of training graphs $\bG^\text{tr}$ and a testing graph $G \in \bG^\text{te}$, our goal is to predict the categories of unlabeled nodes in $G$. This is known as inductive node classification, as we attempt to distill the training graphs to enable node classification in new testing graphs that have not been seen before.

\stitle{Distinction  from few-shot classification.}
While we address the semi-supervised node classification problem, it is worth noting that many meta-learning works \cite{liu2019learning,yao2019graph,zhou2019meta} address the few-shot classification problem. Both problems contain labeled and unlabeled nodes (respectively known as the support and query nodes in the few-shot setting), and thus they may appear similar. However, there are two significant differences.
First, in few-shot classification, the query nodes belong to the same category as at least one of the support nodes. This is often unrealistic on a small graph where some categories only contain one node. In contrast, in our setting, the labeled and unlabeled nodes can be randomly split on any graph. Second, few-shot classification typically deals with novel categories on the same graph, but our setting deals with novel graphs with the same set of categories. 

\subsection{Graph neural networks}\label{sec:prelim:gnn}

Our approach is grounded on graph neural networks, which are inductive due to the shared feature space and weights across graphs. We give a brief review of GNNs in the following.

Modern GNNs generally follow a message passing scheme: each node in a graph receives, maps, and aggregates messages from its neighbors recursively in multiple layers. Specifically, in each layer,
\begin{align} \label{eq.gnn}
\vec{h}^{l+1}_v = \bM(\vec{h}^{l}_v, \{\vec{h}^{l}_u, \forall u \in\bN_v\}; \vec{W}^l),
\end{align}
where $\vec{h}^l_v\in \mathbb{R}^{d_l}$ is the message or the $d_l$-dimensional embedding vector of node $v$ in the $l$-th layer,
$\bN_v$ is the set of neighboring nodes of $v$, $\vec{W}^l\in \mathbb{R}^{d_{l+1}\times d_{l}}$ is a learnable weight matrix to map the node embeddings in the $l$-th layer, and $\bM(\cdot)$ is the message aggregation function.
The initial message of node $v$ in the input layer is simply the original node features, \ie, $\vec{h}^1_v\equiv\vec{x}_v$.
For node classification, the dimension of the output layer is set to the number of node categories and uses a softmax activation.  

The choice of the message aggregation function $\bM$ varies and characterizes different GNN architectures, ranging from a simple mean pooling  \cite{kipf2016semi,hamilton2017inductive,wu2019simplifying} to more complex mechanisms \cite{velivckovic2017graph,hamilton2017inductive,xu2018powerful}. Our proposed model is flexible in the aggregation functions.

\section{METHODOLOGY}
In this section, we present a novel graph inductive framework called MI-GNN, a meta-inductive model that learns to train a model for every new graph.
In the following, we start with an overview of the framework, before we introduce its components in detail.

\subsection{Overview of MI-GNN}

The overarching philosophy of MI-GNN is to design an inductive approach that can dynamically suit to each new graph, in order to cope with the inter-graph differences. A straightforward approach is to train a model on the training graphs, and further perform a fine-tuning step on a new graph in the testing phase.  However, since the training step is independent of the fine-tuning step, it does not train the model to learn how to fine-tune on unseen graphs. 
In contrast, MI-GNN, hinging on the meta-learning principle, learns a general training procedure so that it knows how to dynamically generate a model  suited to any new graph. We set forth the overall framework of MI-GNN in Figure~\ref{fig:framework}. 

First of all, 
MI-GNN exploits each training graph to simulate the semi-supervised node classification task in testing, as shown in  Figure~\ref{fig:framework}(a). Specifically, we take a training graph 
and split its nodes with known labels into two subsets: the \emph{support} set 
and \emph{query} set, 
following the task-based meta-learning setup \cite{finn2017model}. While in a training graph both the support and query nodes have known category labels, we  regard the support nodes as the only labeled nodes and the query nodes as the unlabeled nodes to simulate the semi-supervised classification process during training. On a testing graph, the labeled and unlabeled nodes naturally form the support and query sets, respectively, where the ultimate goal is to predict the unknown categories of the query nodes. 

Next, on the simulated tasks, we learn a task prior and a graph prior during training. The task prior captures the general knowledge of classifying nodes in a semi-supervised setting, whereas the graph  prior captures the general knowledge of transforming the task prior \wrt~each graph. In other words, our general knowledge allows for dual adaptations at both the graph and task levels. On one hand, the graph prior captures and adapts for macro differences across graphs, as illustrated in Figure~\ref{fig:framework}(b). On the other hand, the task prior captures and adapts for micro differences across tasks conditioned on a graph, as illustrated in Figure~\ref{fig:framework}(c).

\subsection{Graphs and tasks} 

\stitle{Training tasks.} We refer to the upper half of Figure~\ref{fig:framework}(a) for illustration. On a training graph $G_i\tr \in \bG\tr$, we can sample a semi-supervised node classification task by randomly splitting its nodes with known labels into the support set $S_i\tr$ and query set $Q_i\tr$ such that $S_i\tr\cap Q_i\tr=\emptyset$. Specifically, without loss of generality, for the node set with known labels $\{v_{i,k}:1\le k \le m+n\}$ on $G_i\tr$, the support and query sets are given by
\begin{align}
S_i\tr &=\{(v_{i,k},\ell(v_{i,k})): 1 \le k \le m \},\\
Q_i\tr &=\{(v_{i,m+k},\ell(v_{i,m+k})): 1 \le k \le n \},
\end{align}
where $m=|S_i\tr|$ and $n=|Q_i\tr|$ denotes the number of nodes in the support and query sets, respectively. Note that for both support and query nodes, their label mapping $\ell$ is known on training graphs. During training, we mimic the model updating process on the support nodes \wrt~their classification loss, and further mimic the prediction process on the query nodes. In particular, the labels of the query nodes are hidden from the model updating process on support nodes, but are used to validate the predictions on the query nodes in order to optimize the general knowledge.  

\stitle{Testing tasks.} On the other hand, suppose a testing graph $G_j\te \in \bG^\text{te}$ has a node set $\{v_{j,k}:1\le k \le m+n\}$ 
such that nodes are  labeled for $1\le k \le m$ only and unlabeled for $m+1\le k \le m+n$. The support and query sets are then given by
\begin{align}
S_j\te &=\{(v_{j,k},\ell(v_{j,k})): 1 \le k \le m \},\\
Q_j\te &=\{ v_{j,m+k}: 1 \le k \le n \}.
\end{align}
The main difference from the training setting is that, the support set contains all the labeled nodes and the query set contains all the unlabeled nodes, as illustrated in the lower half of Figure~\ref{fig:framework}(a). While the labeled support nodes on a testing graph are used in the same way as in training, 
the unlabeled query nodes are only used for prediction and evaluation.

In the following, for brevity we will omit the superscripts \tr\ and \te\ that distinguishes training and testing counterparts (such as $G_i\tr,S_i\tr,Q_i\tr$ and $G_j\te,S_j\te,Q_j\te$) when there is no ambiguity or we are referring to any graph in general (\ie, regardless of training or testing graphs).

\subsection{Graph-level adaptation}

We first formalize the general knowledge consisting of a task prior and a graph prior, which are the foundation of the graph-level adaptation as illustrated in Figure~\ref{fig:framework}(b).

\stitle{Task and graph priors.}
The task prior $\theta$ is designed for quick adaptation to a new semi-supervised node classification task. Given that GNNs can learn  powerful node representations, our task prior takes the form of a GNN model, \ie,   
\begin{align}
\theta=(\vec{W}^1,\vec{W}^2,...),    
\end{align}
where each $\vec{W}^l$ is a learnable weight matrix to map the messages from the neighbors in the $l$-th layer, as introduced in Section~\ref{sec:prelim:gnn}.

Different from most task-based meta learning \cite{snell2017prototypical,finn2017model}, our tasks are not sampled from an i.i.d.~distribution. Instead, tasks are sampled from different graphs, and each task is contextualized and thus conditioned on a graph. We employ a graph prior $\phi$ to condition the task prior, so that the task prior can be transformed to suit each graph. The transformation model is given by 
\begin{align}
\tau(\theta,\vec{g};\phi),
\end{align} 
which (1) is parameterized by the graph prior $\phi$; (2) takes in the task prior $\theta$, and the graph-level representation $\vec{g}$ of an input graph $G$ (which can be either a training or testing graph); (3) outputs a transformed, graph $G$-conditioned task prior. In other words, the graph prior does not directly specify the transformation, but it encodes the rules of how to transform \wrt~each graph. This is essentially a form of hypernetwork \cite{ha2016hypernetworks,perez2018film}, where the task prior is adjusted by a secondary network (parameterized by $\phi$) in response to the changing input graph. 

In the following, we discuss the concrete formulation of the graph-level representation $\vec{g}$, the transformation model $\tau$ and its parameters $\phi$ (\ie, the graph prior).

\stitle{Graph-conditioned transformation.}
To transform the task prior conditioned on a given graph $G$, we need a graph-level representation $\vec{g}$ of the graph. A straightforward approach is to perform a mean pooling of the features or embeddings of all nodes. Although simple, mean pooling does not differentiate the relative importance of each node to the global representation $\vec{g}$. Thus, we adopt an attention-based aggregation to compute our graph-level representation \cite{bai2019unsupervised}, which assigns bigger weights to more significant nodes.

Consider a graph $G_i$ (which can be either a training or testing graph) and its graph-level representation vector $\vec{g}_i$. 
We perform feature-wise linear modulations \cite{perez2018film} on the task prior in order to adapt to $G_i$, by conditioning the transformations on $\vec{g}_i$. This is more flexible than gating, which can only diminish an input as a function of
the same input, instead of a different conditioning input \cite{perez2018film}. 
To be more specific, we use MLPs to generate the scaling vector $\gamma_{i}$ and shifting vector ${\beta}_{i}$ given the input $\vec{g}_i$, which will be used to transform the task prior in order to suit $G_i$. Specifically, 
\begin{align}
\gamma_{i} & = \textsc{MLP}_{\gamma}(\vec{g}_i; \phi_{\gamma}),\label{eq.gamma}\\
\beta_{i} & = \textsc{MLP}_{\beta}(\vec{g}_i; \phi_{\beta}),\label{eq.beta}
\end{align}
where $\phi_{\gamma}$ and $\phi_{\beta}$ are the learnable parameters of the two MLPs, respectively.
Here $\gamma_{i}\in \mathbb{R}^{d_\theta}$ and $\beta_{i}\in \mathbb{R}^{d_\theta}$ are $d_\theta$-dimensional vectors, where $d_\theta$ is the number of parameters in task prior $\theta$. Note that $\theta$ contains all the GNN weight matrices, and we flatten it into a $d_\theta$-dimensional vector in a slight abuse of notation. 

Since $\gamma_i$ and $\beta_i$ have the same dimension as $\theta$, we can apply the transformation in an element-wise manner, to produce the graph $G_i$-conditioned task prior as
\begin{equation}
    \theta_{i} = \tau(\theta,\vec{g}_i;\phi )= (\gamma_{i} + \vec{1})\odot\theta + \beta_{i},\label{eq:graph-level-adapt}
\end{equation}
where $\odot$ denotes element-wise multiplication, and $\vec{1}$ is a vector of ones to ensure that the scaling factors are centered around one.
The graph prior $\phi$, which forms the parameters of $\tau$, consists of the parameters of the two MLPs, \ie,
\begin{align}
    \phi=\{\phi_\gamma,\phi_\beta\}.
\end{align}
Note that $\tau$ is a function of $\vec{g}_i$ as well, since $\gamma_{i}$ and $\beta_{i}$ are functions of $\vec{g}_i$ generated by the two MLPs in Eqs.~\eqref{eq.gamma}--\eqref{eq.beta} in response to the changing input graph. In particular, the two MLPs play the role of secondary networks in the hypernetwork setting \cite{ha2016hypernetworks}.

\subsection{Task-level adaptation}\label{sec:approach:task-adaptation}

Given any graph $G_i$ (training or testing), the graph-conditioned task prior $\theta_i$ serves as a good initialization of the GNN on $G_i$, which can be rapidly adapted to different semi-supervised node classification tasks on $G_i$. Following MAML \cite{finn2017model}, we perform a few gradient descent updates on the support nodes $S_i$ for rapid adaptation, and finally obtain the dual-adapted model $\theta'_i$ as shown in Figure~\ref{fig:framework}(c). Too many updates may cause overfitting to the support nodes and thus hurt the generalization to query nodes, especially when the support set is small in the semi-supervised setting.

The following Eq.~\eqref{eq:one-step} demonstrates one gradient update on the support set $S_i$ \wrt~the graph $G_i$-conditioned $\theta_i$, and extension to multiple steps is straightforward.
\begin{align}
    \theta'_{i}=\theta_{i}-\alpha \frac{\partial L(S_i,\theta_i)}{\partial \theta_{i}},\label{eq:one-step}
\end{align}
where $\alpha \in \mathbb{R}$ is the learning rate of the task-level adaptation, and $L(S_i,\theta_i)$ is the cross-entropy classification loss on the support $S_i$ using the GNN model parameterized by $\theta_i$, as follows.
\begin{align}
\hspace{-1mm}L(S_i,\theta_i)=-\sum_{(v_{i,k},\ell(v_{i,k}))\in S_i} \sum_{c\in C} I(\ell(v_{i,k})=c)\log f(v_{i,k};\theta_i)[c],\label{eq:support-loss}\hspace{-1mm}
\end{align}
where $I(*)$ is an indicator function, $f(*;\theta_i)\in \mathbb{R}^{|C|}$ is the output layer of the GNN parameterized by $\theta_i$ with a softmax activation, and $f(*;\theta_i)[c]$ denotes the probability of category $c$.

\subsection{Overall algorithm}

Finally, we present the algorithm for training and testing. 

\stitle{Training.} Consider a training graph $G_i \in G^\text{tr}$ with graph-level representation $\vec{g}_i$, and a corresponding task $(S_i,Q_i)$. The goal is to optimize the general knowledge in terms of the task prior $\theta$ and graph prior $\phi$ via backpropagation \wrt~the loss on the query nodes after dual adaptions. Specifically, the optimal $\{\theta,\phi\}$ is given by
\begin{align}
\arg\min_{\theta, \phi} \sum_{G_{i} \in \bG\tr}  L(Q_i,\theta_i')+\lambda(\|\gamma_{i}\|_{2}+\|\beta_{i}\|_{2}),\label{eq:overall-optimize}
\end{align}
where (1) $\theta_i'$ is the dual-adapted prior after performing one gradient update according to Eq.~\eqref{eq:one-step} on the $G_i$-condidtioned prior $\theta_i=\tau(\theta, \vec{g}_i;\phi)$, implying that $\theta_i'$ is a function of $\theta$ and $\phi$; (2) $L(Q_i,*)$ is the task loss using the same cross-entropy definition shown in Eq.~\eqref{eq:support-loss}, but computed on the query set $Q_i$; (3) the $L_2$ regularization $\|\gamma_{i}\|_{2}+\|\beta_{i}\|_{2}$ ensures that the scaling is close to 1 and the shifting is close to 0 to prevent overfitting to the training graphs, and $\lambda>0$ is a hyperparameter to control the regularizer. 

In practical implementation,  the optimization is performed over batches of training graphs using any gradient-based optimizer. 
The overall training procedure is outlined in Algorithm~\ref{alg.train}.

\begin{algorithm}[t]
\small
\caption{\textsc{TrainingProcedure}}
\label{alg.train}
\begin{algorithmic}[1]
    \Require training graph set $ \bG^\text{tr}$. 
    \Ensure task prior $\theta$, graph prior $\phi$.
    \State $\theta,\phi\gets$ parameters initialization;
    \While{not converged}
        \State sample a batch of graphs from $\bG^\text{tr}$;
        \For{each graph $G_i$ in the batch}
            \State sample support set $S_i$, query set $Q_i$ from $G_i$;
            \State calculate scaling and shifting factors $\gamma_{i}, \beta_{i}$;
            \Comment{Eqs.~\eqref{eq.gamma},~\eqref{eq.beta}}
            \State $\theta_{i} \gets$ graph-level adaptation on $\theta$; \Comment{Eq.~\eqref{eq:graph-level-adapt}}
            \State calculate support loss $L(S_i,\theta_i)$ and gradient;             \Comment{Eq.~\eqref{eq:support-loss}}
            \State $\theta'_{i} \gets$ task-level adaptation on $\theta_i$;           \Comment{Eq.~\eqref{eq:one-step}}
            \State calculate task (query) loss $L(Q_i,\theta'_{i})$;            
        \EndFor 
        \State $\theta,\phi\gets$ backpropagation of total task loss  \Comment{Eq.~\eqref{eq:overall-optimize}}
    \EndWhile 
    \State \Return $\theta$, $\phi$.
\end{algorithmic}
\end{algorithm}

\stitle{Testing.}
During testing, we follow the same dual adaption mechanism on each testing graph $G_j \in \bG\te$ to generate the dual-adapted prior $\theta_j'$. The only difference from training is that, the query nodes are used for prediction and evaluation, not for backpropagation. That is, for any unlabeled node in the query set $v_{j,k} \in Q_j$, we predict its label as
    $\arg\max_{c\in C} f(v_{j,k};\theta_j')[c]$.

\section{EXPERIMENTS}
In this section, we conduct extensive experiments to evaluate MI-GNN.
More specifically, we compare MI-GNN with state-of-the-art baselines, study the effectiveness of our dual adaptations, and further analyze hyperparameter sensitivity and performance patterns.

\begin{table}[tbp]
    \small
	\centering  
	\caption{Statistics of graph datasets.}  
	\label{table.datasets}  
	\vspace{-2mm}
	\begin{tabular}{@{}c|ccccc@{}}  
		\toprule  
		Dataset&Flickr&Yelp&Cuneiform&COX2&DHFR\\  \midrule
		\# Graphs& 800&800&267& 467& 756\\
		\# Edges (avg.)&13.1&43.5 &20.1&44.8&44.5\\
		\# Nodes (avg.)& 12.5& 6.9 & 21.3 & 41.2 & 42.4 \\
		\# Node features& 500 &300& 3& 3& 3\\
		\# Node classes& 7& 10 &7 & 8& 9\\
		Multi-label? & No & Yes & Yes & No & No\\
		\bottomrule
	\end{tabular}
\end{table}

\subsection{Experimental setup}
\stitle{Datasets.} We conduct experiments on five public graph collections, as follows.
Their statistics are summarized in Table~\ref{table.datasets}. 
\begin{itemize}[leftmargin=*]
    \item Flickr \cite{Zeng2020GraphSAINT:} is a collection of 800 ego-networks sampled from an online image-sharing network. Each node is an image, and each edge connects two images that share some common properties (\eg, same geographic location or gallery).  Our task is to classify each image into one of the seven categories.
    \item Yelp \cite{Zeng2020GraphSAINT:} is a collection of 800 ego-networks sampled from an online review network. Each node represents a user, and each edge represents the friendship relations between users.
    Our task is to classify each user node according to the types of business reviewed by the user in a multi-label setting.
    \item Cuneiform \cite{kriege2018recognizing} is a collection of 267 cuneiform signs in the form of wedge-shaped marks. Each node is a wedge, and each edge indicates the arrangement of the wedges. Our task is to classify the visual appearance of the wedges in a muli-label setting.
   \item COX2 and DHFR \cite{sutherland2003spline} are two collections of molecular structures. Specifically,
     COX2 is a set of 467 cyclooxygenase-2 inhibitors; DHFR is a set of 756 dihydrofolate reductase inhibitors. Each node is an atom and each edge is a chemical bond between two atoms. Our task is to predict the node atomic type.
\end{itemize}

\begin{table*}[tbp]
    \small
	\centering 
	\addtolength{\tabcolsep}{-.5pt}
	\caption{Performance of MI-GNN and baselines, in percent, with 95\% confidence intervals. 
	} 
	\label{table3} 
	\vspace{-2mm} 
    {\footnotesize In each column, the best result is \textbf{bolded} and the runner-up is \underline{underlined}. Improvement by MI-GNN is calculated relative to the best baseline.\\ ***/**/* denotes the difference between MI-GNN and the best baseline is statistically significant at the 0.01/0.05/0.1 level under the two-tail $t$-test.}
    \\[2mm] 
	\begin{tabular}{@{}c|cc|cc|cc|cc|cc@{}}  
		\toprule
		  &\multicolumn{2}{c|}{Flickr}&\multicolumn{2}{c|}{Yelp}&\multicolumn{2}{c|}{Cuneiform}&\multicolumn{2}{c|}{COX2}&\multicolumn{2}{c}{DHFR}\\\cmidrule{2-11}
		  & Accuracy&Micro-F1&Accuracy&Micro-F1&Accuracy&Micro-F1&Accuracy&Micro-F1&Accuracy&Micro-F1
		 \\\midrule
		DeepWalk &39.88$\pm$2.42&30.01$\pm$1.21  & 63.27$\pm$2.73 &57.11$\pm$6.29  & 74.61$\pm$0.60 & 27.05$\pm$2.11 &  37.68$\pm$0.73 & 26.16$\pm$1.08& 33.14$\pm$0.18 & \underline{29.93}$\pm$0.58 \\
		Transduct-GNN &  13.61$\pm$1.22 & 10.71$\pm$1.20 & 24.87$\pm$15.4 &23.85$\pm$14.6 &49.63$\pm$0.95 & 34.00$\pm$1.15  & 13.23$\pm$0.17& \ps{}9.73$\pm$0.22 & 11.21$\pm$0.33 & \ps{}8.65$\pm$0.22 \\
		\midrule
		Planetoid &  14.78$\pm$8.75 & \ps{}8.72$\pm$3.07 & 53.12$\pm$2.38 &46.29$\pm$3.55 &53.14$\pm$5.49 & 30.22$\pm$5.83 & 11.81$\pm$7.41 & 10.58$\pm$8.79 & 17.35$\pm$11.1 & \ps{}9.62$\pm$9.63 \\
		Induct-GNN & 40.48$\pm$1.69 & 29.67$\pm$1.77 & 65.95$\pm$0.56 &56.61$\pm$1.81 &74.89$\pm$0.35 & 18.03$\pm$0.93  & 53.71$\pm$0.92& 41.56$\pm$1.90 & \underline{45.23}$\pm$0.62 &29.38$\pm$6.07\\
		K-NN &   34.11$\pm$1.76 & 26.39$\pm$1.39 & 61.70$\pm$0.90 &57.35$\pm$1.42 &70.36$\pm$0.27 & 35.66$\pm$0.84  & 33.16$\pm$0.95 & 32.84$\pm$1.00 & 36.32$\pm$0.89  & 27.12$\pm$1.20\\
		AGF  &  \underline{40.58}$\pm$1.61 & 28.99$\pm$2.09 & 65.96$\pm$0.54 &56.64$\pm$1.83 &74.89$\pm$0.37 & 18.00$\pm$0.94 & \underline{53.97}$\pm$0.79 & \underline{42.00}$\pm$1.62 & 44.85$\pm$0.56 & 29.08$\pm$5.96\\
		\midrule
		GFL &  30.24$\pm$0.68  & 29.51$\pm$0.69 & 61.62$\pm$0.97& \underline{58.88}$\pm$2.03 &63.72$\pm$0.37 & \underline{38.30}$\pm$0.84  & 29.25$\pm$0.73 & 25.53$\pm$0.94 & 30.24$\pm$0.68 & 29.51$\pm$0.69  \\
		Meta-GNN  &  39.66$\pm$0.92	& \underline{30.02}$\pm$2.49 & \underline{66.24}$\pm$0.84 &56.20$\pm$1.81 &\underline{75.12}$\pm$0.33 & 19.21$\pm$1.25  & 53.24$\pm$0.77 & 37.36$\pm$3.02 & \textbf{45.61}$\pm$0.65 & 28.34$\pm$4.46 \\
		\midrule
		MI-GNN & \textbf{44.45}$\pm$2.18 &\textbf{33.79}$\pm$1.87& \textbf{67.92}$\pm$0.69 &\textbf{60.20}$\pm$2.23&\textbf{81.48}$\pm$0.47&\textbf{43.32}$\pm$1.49 & \textbf{57.27}$\pm$0.80 &\textbf{44.66}$\pm$2.01 & 45.19$\pm$0.70 & \textbf{49.93}$\pm$1.62 \\
		(improv.) & (+9.53\%)&(+12.57\%)&(+2.54\%)&(+2.23\%)&(+8.47\%)&(+13.10\%)&(+6.11\%)&(+6.34\%)&(-0.92\%)&(+66.82\%)\\
		&  ** & ** & *** & *** &*** & ***  & *** & * & & ***\\
	\bottomrule
	\end{tabular}
\end{table*}

\begin{table*}[htbp]
    \small
    \centering  
	\caption{Accuracy of MI-GNN and baselines using alternative GNN architectures, in percent, with 95\% confidence intervals.} 
	\label{table4}  
	\vspace{-2mm} 
	\addtolength{\tabcolsep}{-0.5pt}
	\begin{tabular}{@{}c|ccccc|ccccc@{}}  
	    \toprule 
		& \multicolumn{5}{c|}{\textbf{GCN} as the GNN Architecture} & \multicolumn{5}{c}{\textbf{GraphSAGE} as the GNN Architecture}  \\  
		\cmidrule{2-11}
		&Flickr&Yelp&Cuneiform&COX2&DHFR &Flickr&Yelp&Cuneiform&COX2&DHFR\\  
		\midrule
		Transduct-GNN & 14.89$\pm$0.94 & 50.92$\pm$0.95 & 49.40$\pm$2.27 &  11.89$\pm$0.63 & 10.89$\pm$0.43 & 14.97$\pm$1.96& 50.14$\pm$1.19&50.59$\pm$1.37 & 12.78$\pm$0.65& 11.19$\pm$0.75\\
		Induct-GNN & 12.08$\pm$3.98 & \underline{55.04}$\pm$1.77 & 71.65$\pm$0.46 &86.06$\pm$2.78& \underline{90.31}$\pm$1.03 & \ps 7.31$\pm$1.57& 56.48$\pm$1.73&  84.46$\pm$2.68& 85.28$\pm$1.78& \underline{88.65}$\pm$4.79\\
	    AGF & 11.94$\pm$2.45 & 53.66$\pm$3.04 & 71.66$\pm$0.46  &86.32$\pm$3.08 & 89.64$\pm$1.00 & \ps 7.45$\pm$1.31& 56.70$\pm$2.04&\underline{84.66}$\pm$2.73 & 85.21$\pm$1.85& 88.21$\pm$4.45\\
		Meta-GNN & \underline{22.51}$\pm$3.05 &54.80$\pm$1.86 & \underline{72.24}$\pm$0.88 &\underline{86.92}$\pm$3.66& 90.26$\pm$0.91 & \underline{33.88}$\pm$2.91& \underline{61.80}$\pm$1.81& 84.46$\pm$2.44&  \underline{86.05}$\pm$2.80& 88.17$\pm$4.71\\
		MI-GNN & \textbf{29.91}$\pm$6.85 & \textbf{57.22}$\pm$1.79 &  \textbf{75.36}$\pm$2.07  &\textbf{86.97}$\pm$2.94& \textbf{91.39}$\pm$0.51 & \textbf{42.37}$\pm$3.87& \textbf{69.23}$\pm$1.18& \textbf{91.09}$\pm$2.51& \textbf{93.24}$\pm$0.80& \textbf{93.89}$\pm$0.83 \\
		\bottomrule
	\end{tabular}
\end{table*}

\stitle{Training and testing.} 
For each graph collection, we randomly partition the graphs into 60\%, 20\% and 20\% subsets for training, validation and testing, respectively. On each graph, we randomly split its nodes into two equal halves as the support and query sets, respectively. 
Our goal is to evaluate the performance of node classification on the unlabeled query nodes on the testing graphs, in terms of accuracy and micro-F1. Note that on multi-label graphs with $|C|$ categories, we perform $|C|$ binary classification tasks, one for each category.
Each model is trained with 10 random initializations, and we report the average accuracy and micro-F1 over the 10 runs with 95\% confidence intervals.

\stitle{Settings of MI-GNN.} First, our approach can work with different GNN architectures. By default, we use simplifying graph convolutional networks (SGC)  \cite{wu2019simplifying} in all of our experiments, except in Section~\ref{sec:expt:architecture} where we also adopt GCN \cite{kipf2016semi} and GraphSAGE \cite{hamilton2017inductive} to evaluate the flexibility of MI-GNN. For all GNNs, we employ two layers with a hidden dimension of 16. 
For GraphSAGE, we use the mean aggregator.

Next, for graph-level adaptations, in Eqs.~\eqref{eq.gamma} and \eqref{eq.beta} we adopt MLPs with one hidden layer using LeakyReLU as the activation function, and a linear output layer. For task-level adaptations, we set the number of gradient descent updates to two, and the learning rate of task adaptation $\alpha$ in Eq.~\eqref{eq:one-step} to 0.5 for Flickr, Yelp and Cuneiform or 0.005 for COX2 and DHFR. Lastly, for the overall optimization in Eq.~\eqref{eq:overall-optimize}, we use the Adam optimizer with the learning rate 0.01, and set the regularization co-efficient $\lambda$ to 1 on Flickr and 0.001 on all other datasets. The settings are tuned using the validation graphs.

\stitle{Baselines and settings.} 
We compare our proposed MI-GNN with a comprehensive suite of competitive baselines from three categories.

\vspace{1mm}

\noindent (1) \emph{Transductive} approaches, which do not utilize training graphs. Instead, they directly train the model using the labeled nodes on each testing graph, and we evaluate their classification performance on the unlabeled nodes in the same graph.
\begin{itemize}[leftmargin=*]
    \item DeepWalk \cite{perozzi2014deepwalk}: an unsupervised network embedding method that learns node representations based on the skip-gram model \cite{mikolov2013distributed} to encode random walk sequences. After obtaining node representations on a testing graph, we further train a logistic regression classifier using the labeled nodes. 
    \item Transduct-GNN: applying a GNN in a transductive setting, where it is directly trained on each testing graph.
\end{itemize}
\vspace{1mm}

\noindent (2) \emph{Inductive approaches}, which utilize the training graphs to learn an inductive model that can be applied to new testing graphs. In particular, a fixed inductive model is trained with either no or limited adaptation to the testing graphs.  
\begin{itemize}[leftmargin=*]
    \item Planetoid \cite{yang2016revisiting}: Planetoid is a semi-supervised graph embedding approach. We use its inductive variant in our experiments.
    \item Induct-GNN: applying a GNN in an inductive setting, where it is trained on the training graphs, followed by applying the trained model on each testing graph to generate the node representations. The labeled nodes on the testing graphs are not utilized to adapt the trained model. 
    \item K-NN \cite{triantafillou2019meta}: a two-stage process, in which the first stage is the same as Inductive-GNN, and the second stage subsequently employs a K-nearest-neighbor (K-NN) classifier to classify each unlabeled node into the same category as the closest labeled node in terms of their representations.
    \item AGF \cite{triantafillou2019meta}:  also a two-stage process similar to K-NN, except that in the second stage the K-NN classifier is substituted by a fine-tuning step performed on the labeled nodes.
\end{itemize}
\vspace{1mm}

\noindent (3) \emph{Meta-learning approaches}, which ``learns to learn'' on the training graphs. Instead of learning a fixed model, they learn different forms of general knowledge that can be conveniently adapted to the semi-supervised task on the testing graphs.
\begin{itemize}[leftmargin=*]
    \item GFL \cite{yao2019graph}: a few-shot node classification method on graphs, based on protonets \cite{snell2017prototypical}. 
    While there are major differences between the few-shot and semi-supervised tasks, GFL can still be used in our setting although its performance may not be ideal. 
    \item Meta-GNN \cite{zhou2019meta}: another few-shot node classification approach on graphs, based on MAML \cite{finn2017model}.
\end{itemize}

All methods (except DeepWalk and Planetoid) use the same GNN architecture and corresponding settings in our model. For K-NN, we use the Euclidean distance and set the number of nearest neighbors to 1. For AGF, GFL and Meta-GNN, we use a learning rate of 0.01. For the fine-tuning step in AGF and the task adaptation in Meta-GNN, we use the same setup as the task adaptation in MI-GNN.
For DeepWalk and Planetoid, we set their random walk sampling parameters, such as number of walks, walk length and window size according to their recommended settings, respectively.

\subsection{Performance comparison to baselines}

In Table~\ref{table3},
we report the performance comparison of our proposed MI-GNN and the baselines. 
Generally, our method achieves consistently the best performance among all methods, demonstrating its advantages in inductive semi-supervised node classification. More specifically, we make the following observations.

First, in the transductive setting, Transduct-GNN performs worse than DeepWalk, which is not surprising given that GNNs generally require a large training set to learn effective representations. However, in our setting, an individual test graph may be small with a limited number of labeled nodes. In this regard, the unsupervised representation learning in DeepWalk is more advantageous.

Second, the inductive approaches Induct-GNN and AGF generally outperform transductive approaches, as inductive methods can make use of the abundant training graphs. While K-NN is extended from Induct-GNN with an additional K-nearest neighbor step during testing, it actually performs worse than Induct-GNN. Recall that in our problem setting, on a new graph some node categories may not have labeled nodes (although they have some labeled examples in the training graphs), which makes K-NN unable to classify any node into those categories. Another interesting observation is that, AGF with an additional fine-tuning step on top of Inductive-GNN is only comparable to or marginally better than Inductive-GNN. That means fine-tuning can be prone to overfitting especially when the labeled data are scarce, and a better solution is to learn adaptable general knowledge through meta-learning.

Third, the meta-learning approaches achieve competitive results. 
GFL and Meta-GNN are often better than inductive approaches, but largely trail behind our approach MI-GNN, as they are designed for few-shot classification and lack the dual-level adaptations. In particular, our proposed MI-GNN outperforms all other methods with statistical significance in all but one case. The only exception is on the highly imbalanced DHFR dataset, where MI-GNN achieves slightly worse accuracy than Meta-GNN at low significance ($p=0.442$) but significantly better Micro-F1. Note that Micro-F1 is regarded as a more indicative metric than accuracy on imbalanced classes.

\subsection{Alternative GNN architectures}\label{sec:expt:architecture}

As MI-GNN is designed to work with different GNN architectures, 
we evaluate its flexibility on two other GNN architectures, namely, GCN and GraphSAGE, in addition to SGC as described in the experimental setup. 
For each architecture, we compare with several representative baselines in Table~\ref{table4}. Similar to using SGC, our approach consistently outperforms tranductive, inductive and meta-learning baselines alike. 
The results demonstrate the robustness of our approach across different GNN architectures.

\subsection{Effect of dual adaptations}

The advantage of our approach MI-GNN stems from the dual adaptations at the graph and task levels. 
To investigate the contribution from each level of adaptation, 
we perform an ablation study on MI-GNN, comparing with the following variants. 
(1) \emph{Fine-tune only}: A standard inductive GNN model without any graph- or task-level adaptation, but there is still a simple fine-tuning step on the testing graphs. This is equivalent to the AGF baseline.
(2) \emph{Graph-level only}: This can be obtained by removing the task-level adaptation from MI-GNN.
(3) \emph{Task-level only}: This can be obtained by removing the graph-level adaptation from MI-GNN.

\begin{figure}
   \subfigure[Accuracy]{
   \centering
   \includegraphics[width=0.47\linewidth]{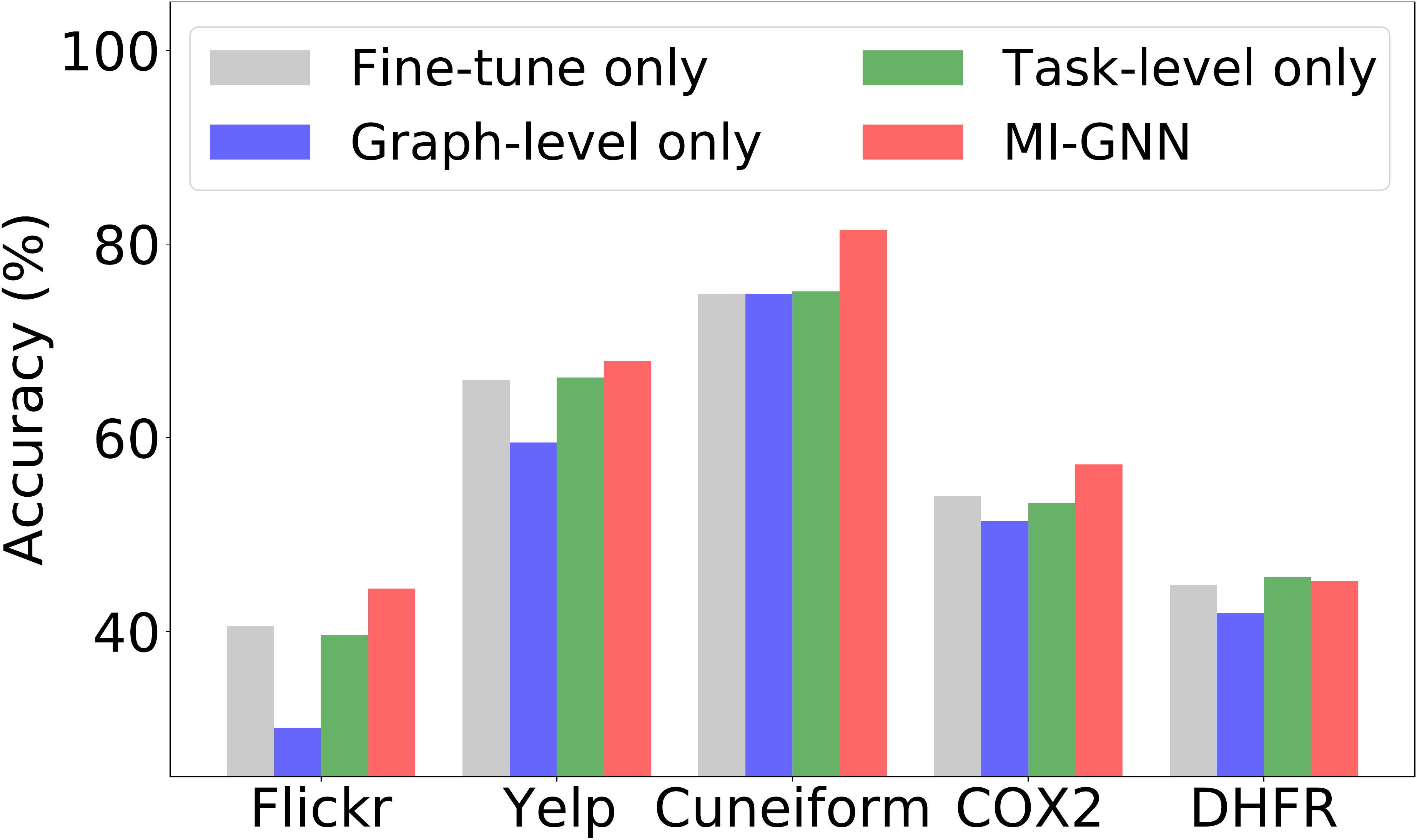}
   }
   \subfigure[Micro-F1]{
   \centering
   \includegraphics[width=0.47\linewidth]{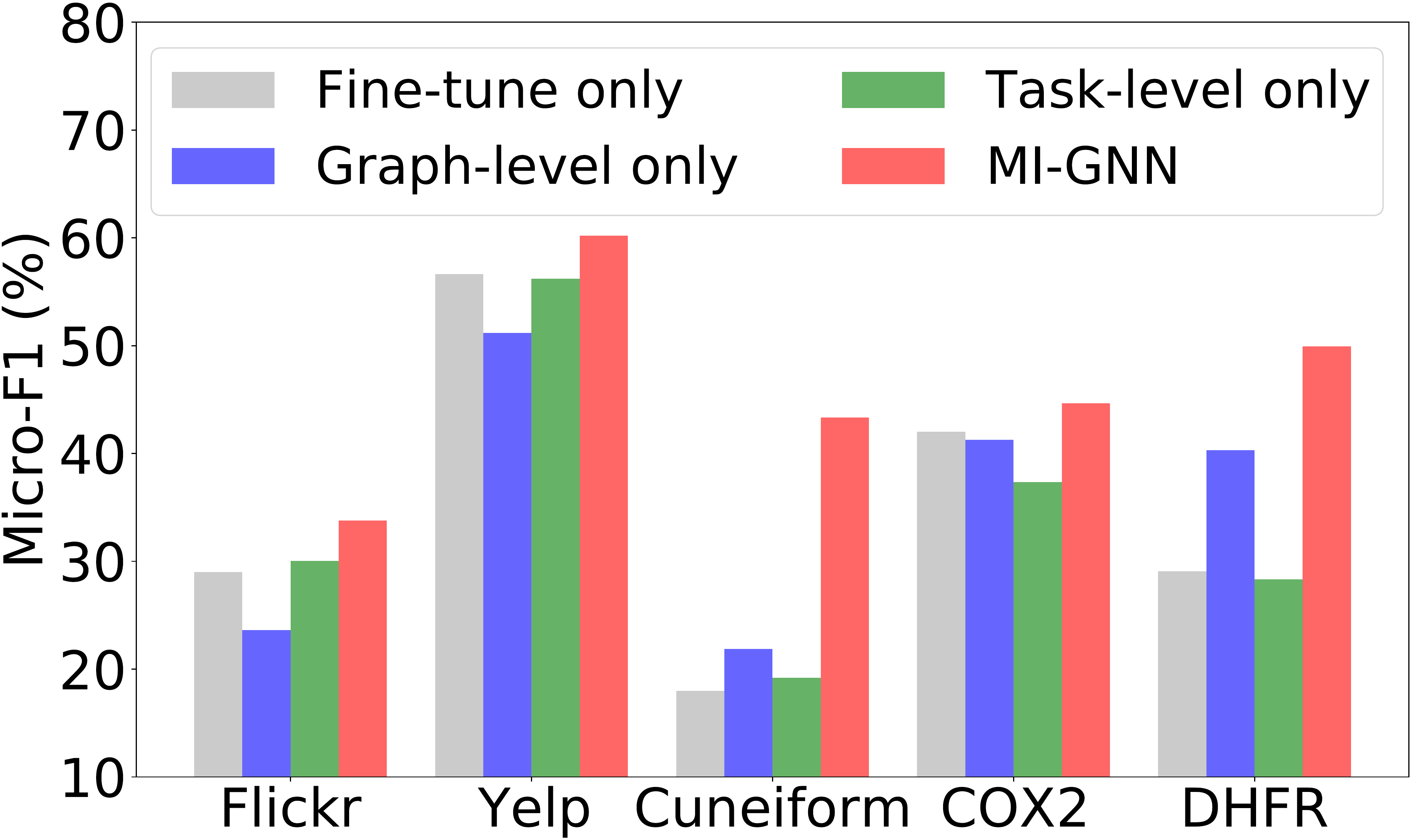}
   }
  \vspace{-4mm}
	\caption{Effect of dual adaptations.}
	\label{fig:ablation}
\end{figure}

We present the comparison in Figure~\ref{fig:ablation}. First of all, MI-GNN outperforms all the ablated models consistently,
demonstrating the overall benefit of the dual adaptations. Among the ablated models, Fine-tune only achieves a surprisingly competitive performance approaching the model with only task-level adaptation, while graph-level adaptation performs rather poorly in majority of the cases. That means in MI-GNN the two levels of adaptations are both crucial and they are well integrated, as each adaptation alone may not give any significant benefit over a simple fine-tuning step but together they work much better.

\subsection{Hyperparameter sensitivity}

We study the effect of regularization in graph-level adaptation, and the number of gradient descent steps in task-level adaptation.

\begin{figure}
    \centering
   \subfigure[Regularization]{
   \centering
   \includegraphics[width=0.47\linewidth]{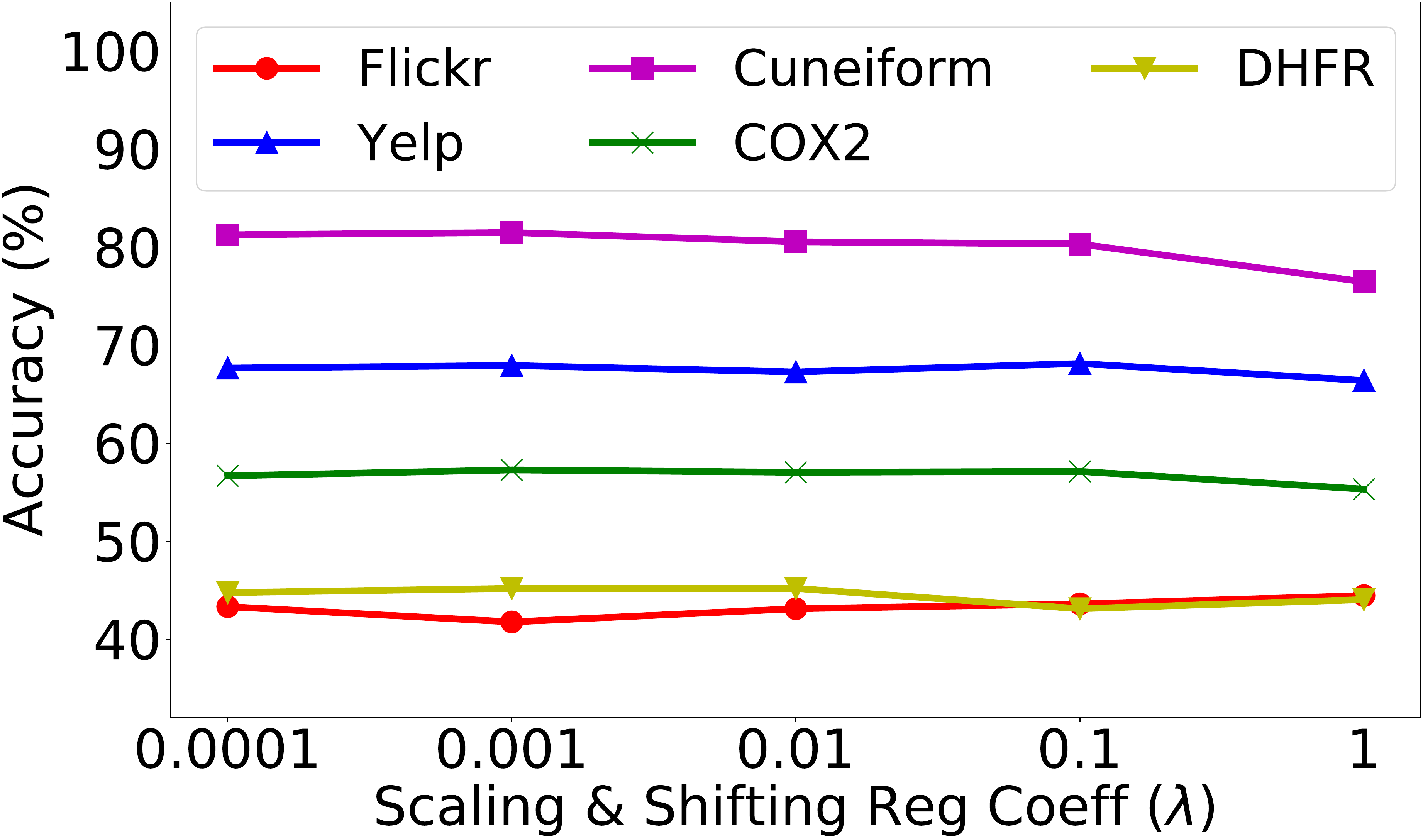}
   }
   \subfigure[Gradient steps]{
   \centering
   \includegraphics[width=0.47\linewidth]{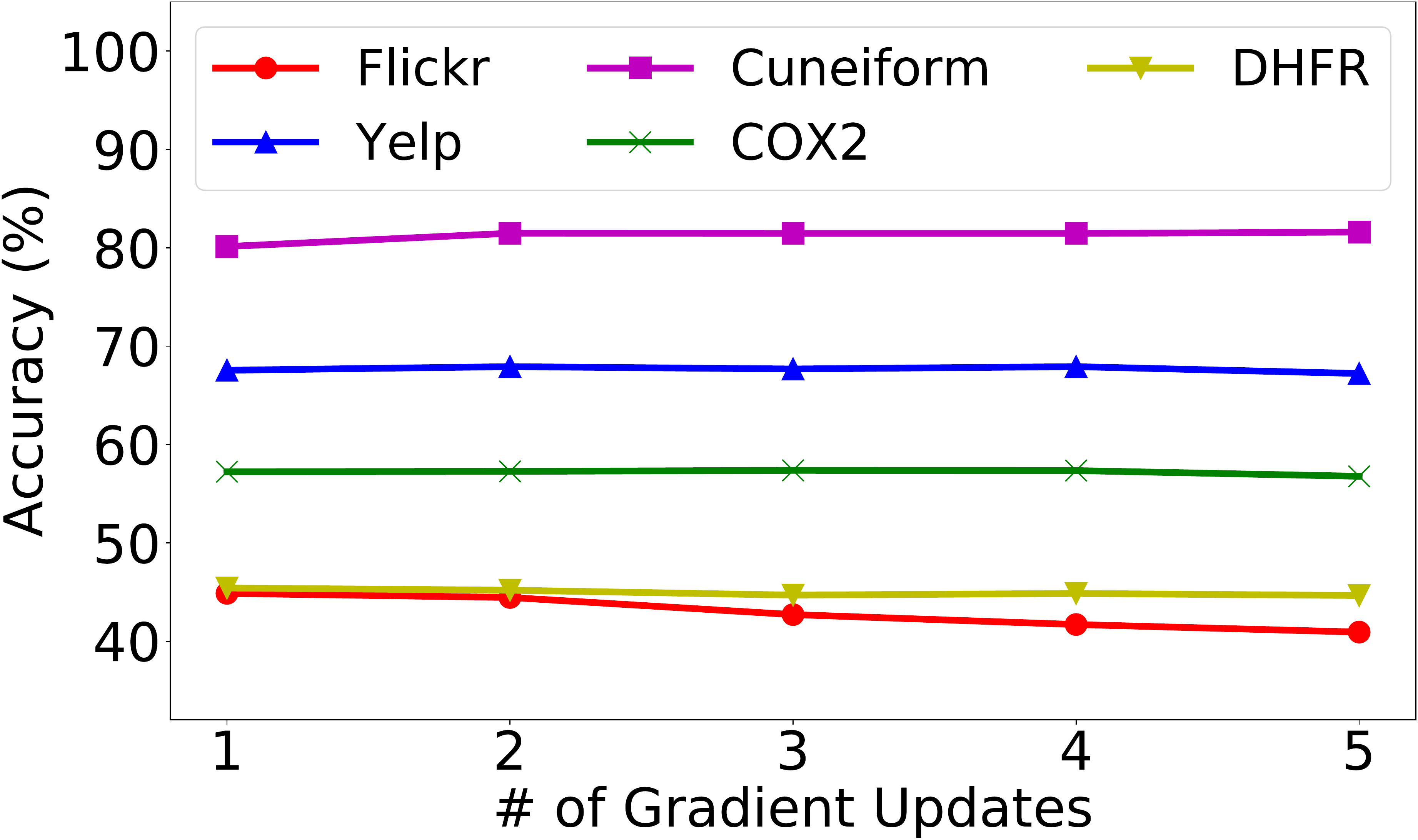}
   }
  \vspace{-4mm}
	\caption{Impact of regularization and gradient steps.}
	\label{fig:parameters}
\end{figure}

\stitle{Regularization for graph-level adaptation.}
To prevent overfitting to the training graphs, we constrain the graph-conditioned transformations to ensure that the scaling is close to 1 and the shifting is close to 0. We study the effect of the regularization in Figure~\ref{fig:parameters}(a), as controlled by the co-efficient $\lambda$ in Eq.~\eqref{eq:overall-optimize}. In general, the performance is stable for different values of $\lambda$, although smaller values in the range $[0.0001, 0.01]$ tends to perform better. 
Overly large values will result in very little scaling and shifting, effectively removing the graph-level adaptation and thus suffering from reduced performance.

\stitle{Number of gradient steps in task-level adaptation.}
As discussed in Section~\ref{sec:approach:task-adaptation}, we achieve task-level adaptation by conducting a few steps of gradient descent on the support set of each graph. To understand the impact of number of steps, we conduct experiments using different number of steps. Results in Figure~\ref{fig:parameters}(b) reveal that the performance is not sensitive to the number of steps. Thus, it is sufficient to  perform just one or two steps for efficiency.

\begin{figure}
	\centering
   \subfigure[Accuracy]{
   \centering
   \includegraphics[width=0.47\linewidth]{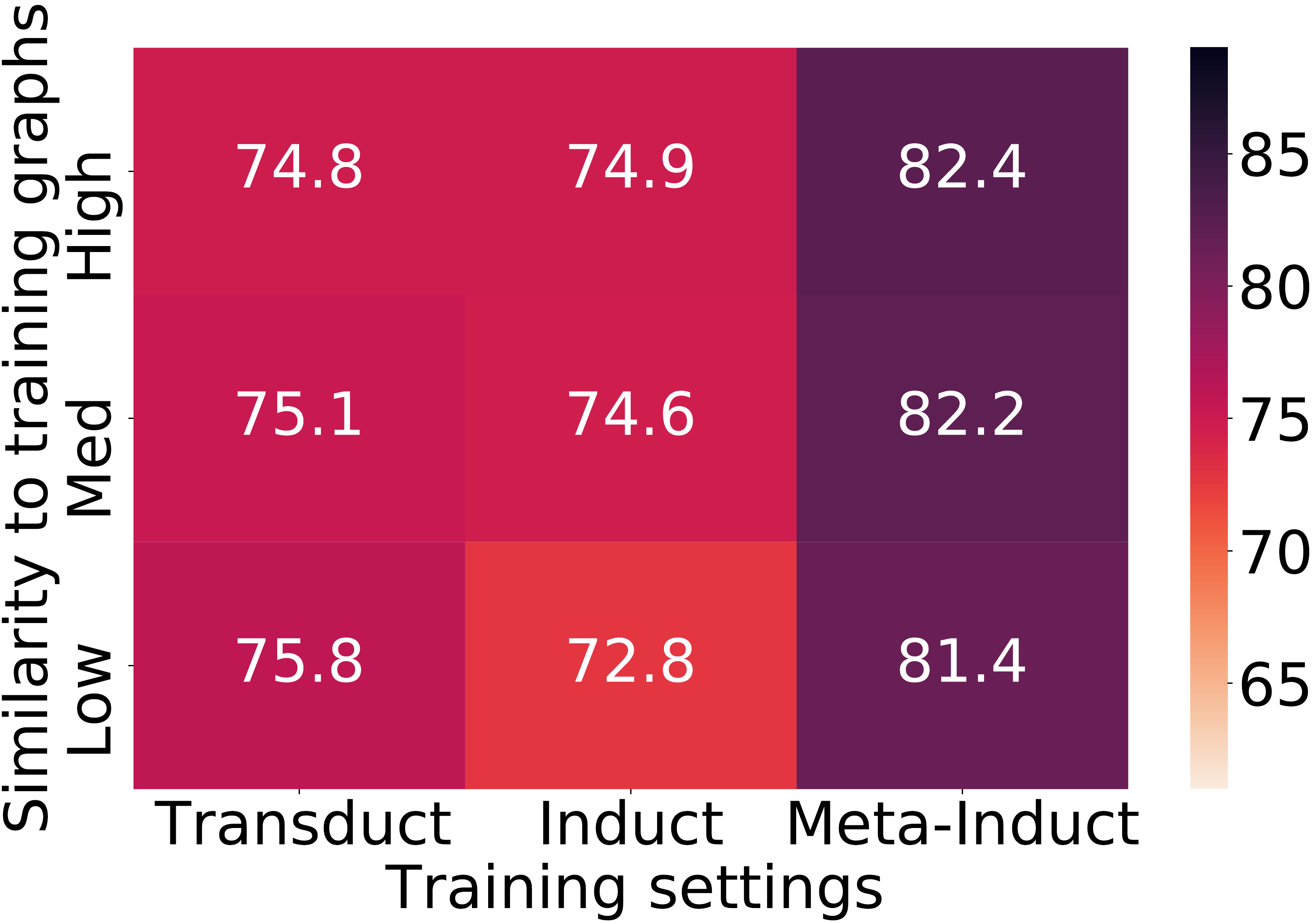}
   }
   \subfigure[Micro-F1]{
   \centering
   \includegraphics[width=0.47\linewidth]{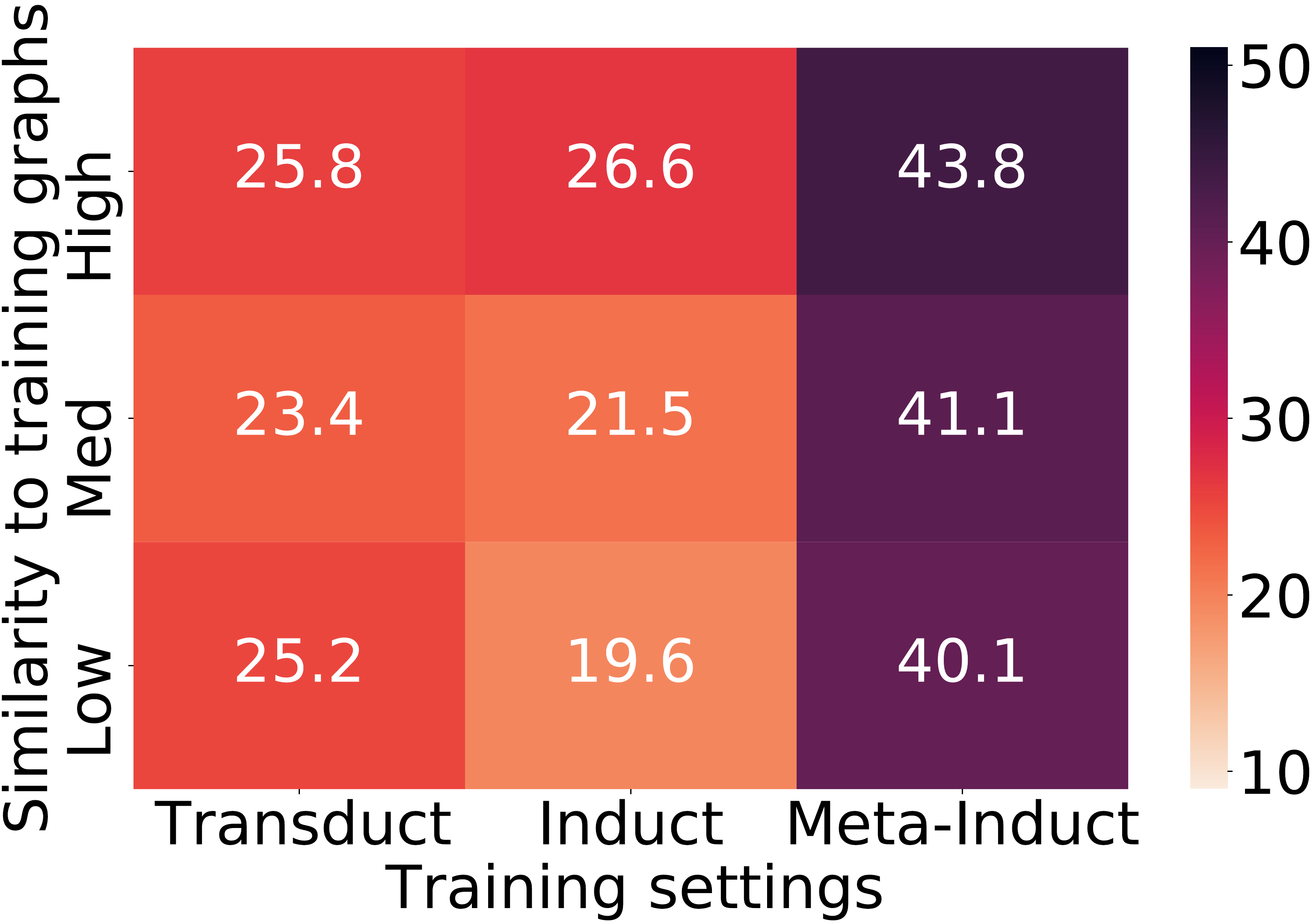}
   }
  \vspace{-4mm}
	\caption{Performance \wrt~similarity to training graphs.}
	\label{fig:negative}
\end{figure}

\subsection{Performance case study}
To understand more precisely when our proposed meta-inductive framework can be effective, we conduct 
a case study on the performance patterns of transductive and inductive methods. 
On one hand, the performance of inductive models on testing graphs would directly correlate to the similarity between testing
and training graphs. Intuitively, the less similar they are, the less effectively knowledge can be transferred from training to testing graphs.
On the other hand, transductive methods are not influenced by such similarity, as they do not learn from training graphs at all.

We compute the similarity between two graphs based on the Euclidean distance of their graph-level representations generated by an attention-based model \cite{bai2019unsupervised}.
The similarity between  a testing graph and a set of training graphs is then given by the average similarity between the testing graph and each of the training graph. Subsequently, we split the testing graphs into three groups according to their similarity to the training set, namely, high, medium and low similarity. We report the performance of each group under three settings: transductive (using DeepWalk), inductive (using Induct-GNN) and meta-inductive (using MI-GNN). 

We present heatmap visualizations in Figure~\ref{fig:negative} on the Cuneiform dataset. Although the inductive setting can leverage knowledge gained from the training graphs and potentially transfer it to testing graphs, it is not always helpful and can even be harmful when the training data are quite different from the testing data, known as negative transfer \cite{rosenstein2005transfer}. 
Our heatmaps show that in the transductive setting, the performance remains largely unchanged across the three groups, as transductive methods do not rely on any knowledge transfer from training graphs. In contrast, the conventional inductive setting can suffer from negative transfer, as its performance drops considerably when the testing graphs become less similar to the training graphs. Finally, our meta-inductive approach is generally robust and the effect of negative transfer is much smaller than the conventional inductive method. The underlying reason is that we only learn a form of general knowledge from training graphs, which undergoes a further adaptation process to suit each testing graph. The adaptation process makes our method more robust when dealing with different graphs, which is also our key distinction from conventional inductive methods.


\section{Conclusion}
In this paper, we studied the problem of inductive node classification across graphs. Unlike existing one-model-fits-all approaches, we proposed a novel framework called MI-GNN to customize the inductive model to each graph under a meta-learning paradigm. To cope with the differences across graphs, we designed a dual adaptation mechanism at both the graph and task levels. More specifically, we learn a graph prior to adapt for the graph-level differences, and a task prior to further adapt for the task-level differences conditioned on each graph. Extensive experiments on five real-world graph collections demonstrate the effectiveness of MI-GNN.


\begin{acks}
This research is supported by the Agency for Science, Technology and Research (A*STAR) under its AME Programmatic Funds (Grant No. A20H6b0151).
\end{acks}

\newpage

\balance

\bibliographystyle{ACM-Reference-Format}
\bibliography{references}


\begin{thebibliography}{57}


\ifx \showCODEN    \undefined \def \showCODEN     #1{\unskip}     \fi
\ifx \showDOI      \undefined \def \showDOI       #1{#1}\fi
\ifx \showISBNx    \undefined \def \showISBNx     #1{\unskip}     \fi
\ifx \showISBNxiii \undefined \def \showISBNxiii  #1{\unskip}     \fi
\ifx \showISSN     \undefined \def \showISSN      #1{\unskip}     \fi
\ifx \showLCCN     \undefined \def \showLCCN      #1{\unskip}     \fi
\ifx \shownote     \undefined \def \shownote      #1{#1}          \fi
\ifx \showarticletitle \undefined \def \showarticletitle #1{#1}   \fi
\ifx \showURL      \undefined \def \showURL       {\relax}        \fi
\providecommand\bibfield[2]{#2}
\providecommand\bibinfo[2]{#2}
\providecommand\natexlab[1]{#1}
\providecommand\showeprint[2][]{arXiv:#2}

\bibitem[\protect\citeauthoryear{Bai, Ding, Qiao, Marinovic, Gu, Chen, Sun, and
  Wang}{Bai et~al\mbox{.}}{2019}]%
        {bai2019unsupervised}
\bibfield{author}{\bibinfo{person}{Yunsheng Bai}, \bibinfo{person}{Hao Ding},
  \bibinfo{person}{Yang Qiao}, \bibinfo{person}{Agustin Marinovic},
  \bibinfo{person}{Ken Gu}, \bibinfo{person}{Ting Chen},
  \bibinfo{person}{Yizhou Sun}, {and} \bibinfo{person}{Wei Wang}.}
  \bibinfo{year}{2019}\natexlab{}.
\newblock \showarticletitle{Unsupervised inductive graph-level representation
  learning via graph-graph proximity}. In \bibinfo{booktitle}{\emph{IJCAI}}.
  \bibinfo{pages}{1988--1994}.
\newblock


\bibitem[\protect\citeauthoryear{Blum and Chawla}{Blum and Chawla}{2001}]%
        {blum_learning_2001}
\bibfield{author}{\bibinfo{person}{Avrim Blum} {and} \bibinfo{person}{Shuchi
  Chawla}.} \bibinfo{year}{2001}\natexlab{}.
\newblock \showarticletitle{Learning from labeled and unlabeled data using
  graph mincuts}. In \bibinfo{booktitle}{\emph{ICML}}. \bibinfo{pages}{19--26}.
\newblock


\bibitem[\protect\citeauthoryear{Brockschmidt}{Brockschmidt}{2020}]%
        {brockschmidt2019gnn}
\bibfield{author}{\bibinfo{person}{Marc Brockschmidt}.}
  \bibinfo{year}{2020}\natexlab{}.
\newblock \showarticletitle{GNN-FiLM: Graph neural networks with feature-wise
  linear modulation}. In \bibinfo{booktitle}{\emph{ICML}}.
  \bibinfo{pages}{1144--1152}.
\newblock


\bibitem[\protect\citeauthoryear{Cai, Zheng, and Chang}{Cai
  et~al\mbox{.}}{2018}]%
        {cai2018comprehensive}
\bibfield{author}{\bibinfo{person}{Hongyun Cai}, \bibinfo{person}{Vincent~W
  Zheng}, {and} \bibinfo{person}{Kevin Chen-Chuan Chang}.}
  \bibinfo{year}{2018}\natexlab{}.
\newblock \showarticletitle{A comprehensive survey of graph embedding:
  Problems, techniques, and applications}.
\newblock \bibinfo{journal}{\emph{TKDE}} \bibinfo{volume}{30},
  \bibinfo{number}{9} (\bibinfo{year}{2018}), \bibinfo{pages}{1616--1637}.
\newblock


\bibitem[\protect\citeauthoryear{Ding, Wang, Li, Shu, Liu, and Liu}{Ding
  et~al\mbox{.}}{2020}]%
        {ding2020graph}
\bibfield{author}{\bibinfo{person}{Kaize Ding}, \bibinfo{person}{Jianling
  Wang}, \bibinfo{person}{Jundong Li}, \bibinfo{person}{Kai Shu},
  \bibinfo{person}{Chenghao Liu}, {and} \bibinfo{person}{Huan Liu}.}
  \bibinfo{year}{2020}\natexlab{}.
\newblock \showarticletitle{Graph prototypical networks for few-shot learning
  on attributed networks}. In \bibinfo{booktitle}{\emph{CIKM}}.
  \bibinfo{pages}{295--304}.
\newblock


\bibitem[\protect\citeauthoryear{Dong, Yuan, Yao, Xu, and Zhu}{Dong
  et~al\mbox{.}}{2020}]%
        {dong2020mamo}
\bibfield{author}{\bibinfo{person}{Manqing Dong}, \bibinfo{person}{Feng Yuan},
  \bibinfo{person}{Lina Yao}, \bibinfo{person}{Xiwei Xu}, {and}
  \bibinfo{person}{Liming Zhu}.} \bibinfo{year}{2020}\natexlab{}.
\newblock \showarticletitle{MAMO: Memory-augmented meta-optimization for
  cold-start recommendation}. In \bibinfo{booktitle}{\emph{KDD}}.
  \bibinfo{pages}{688--697}.
\newblock


\bibitem[\protect\citeauthoryear{Dougnon, Fournier-Viger, Lin, and
  Nkambou}{Dougnon et~al\mbox{.}}{2016}]%
        {dougnon2016inferring}
\bibfield{author}{\bibinfo{person}{Ra{\"\i}ssa~Yapan Dougnon},
  \bibinfo{person}{Philippe Fournier-Viger}, \bibinfo{person}{Jerry Chun-Wei
  Lin}, {and} \bibinfo{person}{Roger Nkambou}.}
  \bibinfo{year}{2016}\natexlab{}.
\newblock \showarticletitle{Inferring social network user profiles using a
  partial social graph}.
\newblock \bibinfo{journal}{\emph{Journal of Intelligent Information Systems}}
  \bibinfo{volume}{47}, \bibinfo{number}{2} (\bibinfo{year}{2016}),
  \bibinfo{pages}{313--344}.
\newblock


\bibitem[\protect\citeauthoryear{Du, Wang, Song, Lu, and Wang}{Du
  et~al\mbox{.}}{2018}]%
        {du2018dynamic}
\bibfield{author}{\bibinfo{person}{Lun Du}, \bibinfo{person}{Yun Wang},
  \bibinfo{person}{Guojie Song}, \bibinfo{person}{Zhicong Lu}, {and}
  \bibinfo{person}{Junshan Wang}.} \bibinfo{year}{2018}\natexlab{}.
\newblock \showarticletitle{Dynamic network embedding: an extended approach for
  skip-gram based network embedding.}. In \bibinfo{booktitle}{\emph{IJCAI}}.
  \bibinfo{pages}{2086--2092}.
\newblock


\bibitem[\protect\citeauthoryear{Fang, Chang, and Lauw}{Fang
  et~al\mbox{.}}{2014}]%
        {DBLP:conf/icml/FangCL14}
\bibfield{author}{\bibinfo{person}{Yuan Fang},
  \bibinfo{person}{Kevin~Chen{-}Chuan Chang}, {and}
  \bibinfo{person}{Hady~Wirawan Lauw}.} \bibinfo{year}{2014}\natexlab{}.
\newblock \showarticletitle{Graph-based semi-supervised learning: Realizing
  pointwise smoothness probabilistically}. In \bibinfo{booktitle}{\emph{ICML}}.
  \bibinfo{pages}{406--414}.
\newblock


\bibitem[\protect\citeauthoryear{Fang, Hsu, and Chang}{Fang
  et~al\mbox{.}}{2012}]%
        {DBLP:conf/sigir/FangHC12}
\bibfield{author}{\bibinfo{person}{Yuan Fang}, \bibinfo{person}{Bo{-}June~Paul
  Hsu}, {and} \bibinfo{person}{Kevin~Chen{-}Chuan Chang}.}
  \bibinfo{year}{2012}\natexlab{}.
\newblock \showarticletitle{Confidence-aware graph regularization with
  heterogeneous pairwise features}. In \bibinfo{booktitle}{\emph{SIGIR}}.
  \bibinfo{pages}{951--960}.
\newblock


\bibitem[\protect\citeauthoryear{Finn, Abbeel, and Levine}{Finn
  et~al\mbox{.}}{2017}]%
        {finn2017model}
\bibfield{author}{\bibinfo{person}{Chelsea Finn}, \bibinfo{person}{Pieter
  Abbeel}, {and} \bibinfo{person}{Sergey Levine}.}
  \bibinfo{year}{2017}\natexlab{}.
\newblock \showarticletitle{Model-agnostic meta-learning for fast adaptation of
  deep networks}.
\newblock \bibinfo{journal}{\emph{ICML}} (\bibinfo{year}{2017}),
  \bibinfo{pages}{1126--1135}.
\newblock


\bibitem[\protect\citeauthoryear{Goyal, Chhetri, and Canedo}{Goyal
  et~al\mbox{.}}{2020}]%
        {goyal2020dyngraph2vec}
\bibfield{author}{\bibinfo{person}{Palash Goyal}, \bibinfo{person}{Sujit~Rokka
  Chhetri}, {and} \bibinfo{person}{Arquimedes Canedo}.}
  \bibinfo{year}{2020}\natexlab{}.
\newblock \showarticletitle{dyngraph2vec: Capturing network dynamics using
  dynamic graph representation learning}.
\newblock \bibinfo{journal}{\emph{Knowledge-Based Systems}}
  \bibinfo{volume}{187} (\bibinfo{year}{2020}), \bibinfo{pages}{104816}.
\newblock


\bibitem[\protect\citeauthoryear{Goyal, Kamra, He, and Liu}{Goyal
  et~al\mbox{.}}{2018}]%
        {goyal2018dyngem}
\bibfield{author}{\bibinfo{person}{Palash Goyal}, \bibinfo{person}{Nitin
  Kamra}, \bibinfo{person}{Xinran He}, {and} \bibinfo{person}{Yan Liu}.}
  \bibinfo{year}{2018}\natexlab{}.
\newblock \showarticletitle{{DynGEM}: Deep embedding method for dynamic
  graphs}.
\newblock \bibinfo{journal}{\emph{arXiv preprint arXiv:1805.11273}}
  (\bibinfo{year}{2018}).
\newblock


\bibitem[\protect\citeauthoryear{Grover and Leskovec}{Grover and
  Leskovec}{2016}]%
        {grover2016node2vec}
\bibfield{author}{\bibinfo{person}{Aditya Grover} {and} \bibinfo{person}{Jure
  Leskovec}.} \bibinfo{year}{2016}\natexlab{}.
\newblock \showarticletitle{node2vec: Scalable feature learning for networks}.
  In \bibinfo{booktitle}{\emph{KDD}}. \bibinfo{pages}{855--864}.
\newblock


\bibitem[\protect\citeauthoryear{Ha, Dai, and Le}{Ha et~al\mbox{.}}{2017}]%
        {ha2016hypernetworks}
\bibfield{author}{\bibinfo{person}{David Ha}, \bibinfo{person}{Andrew Dai},
  {and} \bibinfo{person}{Quoc~V Le}.} \bibinfo{year}{2017}\natexlab{}.
\newblock \showarticletitle{Hypernetworks}. In
  \bibinfo{booktitle}{\emph{ICLR}}.
\newblock


\bibitem[\protect\citeauthoryear{Hamilton, Ying, and Leskovec}{Hamilton
  et~al\mbox{.}}{2017}]%
        {hamilton2017inductive}
\bibfield{author}{\bibinfo{person}{Will Hamilton}, \bibinfo{person}{Zhitao
  Ying}, {and} \bibinfo{person}{Jure Leskovec}.}
  \bibinfo{year}{2017}\natexlab{}.
\newblock \showarticletitle{Inductive representation learning on large graphs}.
  In \bibinfo{booktitle}{\emph{NeurIPS}}. \bibinfo{pages}{1024--1034}.
\newblock


\bibitem[\protect\citeauthoryear{He, Carbonell, and Liu}{He
  et~al\mbox{.}}{2007}]%
        {he2007graph}
\bibfield{author}{\bibinfo{person}{Jingrui He}, \bibinfo{person}{Jaime
  Carbonell}, {and} \bibinfo{person}{Yan Liu}.}
  \bibinfo{year}{2007}\natexlab{}.
\newblock \showarticletitle{Graph-based semi-supervised learning as a
  generative model}. In \bibinfo{booktitle}{\emph{IJCAI}}.
  \bibinfo{pages}{2492--2497}.
\newblock


\bibitem[\protect\citeauthoryear{Hu, Liu, Gomes, Zitnik, Liang, Pande, and
  Leskovec}{Hu et~al\mbox{.}}{2020}]%
        {hu2019strategies}
\bibfield{author}{\bibinfo{person}{Weihua Hu}, \bibinfo{person}{Bowen Liu},
  \bibinfo{person}{Joseph Gomes}, \bibinfo{person}{Marinka Zitnik},
  \bibinfo{person}{Percy Liang}, \bibinfo{person}{Vijay Pande}, {and}
  \bibinfo{person}{Jure Leskovec}.} \bibinfo{year}{2020}\natexlab{}.
\newblock \showarticletitle{Strategies for pre-training graph neural networks}.
  In \bibinfo{booktitle}{\emph{ICLR}}.
\newblock


\bibitem[\protect\citeauthoryear{Hu, Chen, Chang, and Sun}{Hu
  et~al\mbox{.}}{2019}]%
        {hu2019few}
\bibfield{author}{\bibinfo{person}{Ziniu Hu}, \bibinfo{person}{Ting Chen},
  \bibinfo{person}{Kai-Wei Chang}, {and} \bibinfo{person}{Yizhou Sun}.}
  \bibinfo{year}{2019}\natexlab{}.
\newblock \showarticletitle{Few-shot representation learning for
  out-of-vocabulary words}. In \bibinfo{booktitle}{\emph{ACL}}.
  \bibinfo{pages}{4102--4112}.
\newblock


\bibitem[\protect\citeauthoryear{Kipf and Welling}{Kipf and Welling}{2017}]%
        {kipf2016semi}
\bibfield{author}{\bibinfo{person}{Thomas~N Kipf} {and} \bibinfo{person}{Max
  Welling}.} \bibinfo{year}{2017}\natexlab{}.
\newblock \showarticletitle{Semi-supervised classification with graph
  convolutional networks}. In \bibinfo{booktitle}{\emph{ICLR}}.
\newblock


\bibitem[\protect\citeauthoryear{Kriege, Fey, Fisseler, Mutzel, and
  Weichert}{Kriege et~al\mbox{.}}{2018}]%
        {kriege2018recognizing}
\bibfield{author}{\bibinfo{person}{Nils~M Kriege}, \bibinfo{person}{Matthias
  Fey}, \bibinfo{person}{Denis Fisseler}, \bibinfo{person}{Petra Mutzel}, {and}
  \bibinfo{person}{Frank Weichert}.} \bibinfo{year}{2018}\natexlab{}.
\newblock \showarticletitle{Recognizing cuneiform signs using graph based
  methods}. In \bibinfo{booktitle}{\emph{International Workshop on
  Cost-Sensitive Learning}}. \bibinfo{pages}{31--44}.
\newblock


\bibitem[\protect\citeauthoryear{Li, Wang, and Chang}{Li et~al\mbox{.}}{2014}]%
        {li2014user}
\bibfield{author}{\bibinfo{person}{Rui Li}, \bibinfo{person}{Chi Wang}, {and}
  \bibinfo{person}{Kevin Chen-Chuan Chang}.} \bibinfo{year}{2014}\natexlab{}.
\newblock \showarticletitle{User profiling in an ego network: co-profiling
  attributes and relationships}. In \bibinfo{booktitle}{\emph{WWW}}.
  \bibinfo{pages}{819--830}.
\newblock


\bibitem[\protect\citeauthoryear{Liu, Zhou, Long, Jiang, and Zhang}{Liu
  et~al\mbox{.}}{2019}]%
        {liu2019learning}
\bibfield{author}{\bibinfo{person}{Lu Liu}, \bibinfo{person}{Tianyi Zhou},
  \bibinfo{person}{Guodong Long}, \bibinfo{person}{Jing Jiang}, {and}
  \bibinfo{person}{Chengqi Zhang}.} \bibinfo{year}{2019}\natexlab{}.
\newblock \showarticletitle{Learning to propagate for graph meta-learning}. In
  \bibinfo{booktitle}{\emph{NeurIPS}}. \bibinfo{pages}{1039--1050}.
\newblock


\bibitem[\protect\citeauthoryear{Liu, Fang, Liu, and Hoi}{Liu
  et~al\mbox{.}}{2021a}]%
        {liu2021nodewise}
\bibfield{author}{\bibinfo{person}{Zemin Liu}, \bibinfo{person}{Yuan Fang},
  \bibinfo{person}{Chenghao Liu}, {and} \bibinfo{person}{Steven~C.H. Hoi}.}
  \bibinfo{year}{2021}\natexlab{a}.
\newblock \showarticletitle{Node-wise localization of graph neural networks}.
  In \bibinfo{booktitle}{\emph{IJCAI}}.
\newblock


\bibitem[\protect\citeauthoryear{Liu, Fang, Liu, and Hoi}{Liu
  et~al\mbox{.}}{2021b}]%
        {liu2021relative}
\bibfield{author}{\bibinfo{person}{Zemin Liu}, \bibinfo{person}{Yuan Fang},
  \bibinfo{person}{Chenghao Liu}, {and} \bibinfo{person}{Steven~C.H. Hoi}.}
  \bibinfo{year}{2021}\natexlab{b}.
\newblock \showarticletitle{Relative and absolute location embedding for
  few-shot node classification on graph}. In \bibinfo{booktitle}{\emph{AAAI}}.
\newblock


\bibitem[\protect\citeauthoryear{Liu, Zhang, Fang, Zhang, and Hoi}{Liu
  et~al\mbox{.}}{2020}]%
        {liu2020towards}
\bibfield{author}{\bibinfo{person}{Zemin Liu}, \bibinfo{person}{Wentao Zhang},
  \bibinfo{person}{Yuan Fang}, \bibinfo{person}{Xinming Zhang}, {and}
  \bibinfo{person}{Steven~C.H. Hoi}.} \bibinfo{year}{2020}\natexlab{}.
\newblock \showarticletitle{Towards locality-aware meta-learning of tail node
  embeddings on networks}. In \bibinfo{booktitle}{\emph{CIKM}}.
  \bibinfo{pages}{975--984}.
\newblock


\bibitem[\protect\citeauthoryear{Lu, Jiang, Fang, and Shi}{Lu
  et~al\mbox{.}}{2021}]%
        {lu2021learning}
\bibfield{author}{\bibinfo{person}{Yuanfu Lu}, \bibinfo{person}{Xunqiang
  Jiang}, \bibinfo{person}{Yuan Fang}, {and} \bibinfo{person}{Chuan Shi}.}
  \bibinfo{year}{2021}\natexlab{}.
\newblock \showarticletitle{Learning to pre-train graph neural networks}. In
  \bibinfo{booktitle}{\emph{AAAI}}.
\newblock


\bibitem[\protect\citeauthoryear{Mikolov, Sutskever, Chen, Corrado, and
  Dean}{Mikolov et~al\mbox{.}}{2013}]%
        {mikolov2013distributed}
\bibfield{author}{\bibinfo{person}{Tomas Mikolov}, \bibinfo{person}{Ilya
  Sutskever}, \bibinfo{person}{Kai Chen}, \bibinfo{person}{Greg~S Corrado},
  {and} \bibinfo{person}{Jeff Dean}.} \bibinfo{year}{2013}\natexlab{}.
\newblock \showarticletitle{Distributed representations of words and phrases
  and their compositionality}. In \bibinfo{booktitle}{\emph{NeurIPS}}.
  \bibinfo{pages}{3111--3119}.
\newblock


\bibitem[\protect\citeauthoryear{Nguyen, Lee, Rossi, Ahmed, Koh, and
  Kim}{Nguyen et~al\mbox{.}}{2018}]%
        {nguyen2018continuous}
\bibfield{author}{\bibinfo{person}{Giang~Hoang Nguyen},
  \bibinfo{person}{John~Boaz Lee}, \bibinfo{person}{Ryan~A Rossi},
  \bibinfo{person}{Nesreen~K Ahmed}, \bibinfo{person}{Eunyee Koh}, {and}
  \bibinfo{person}{Sungchul Kim}.} \bibinfo{year}{2018}\natexlab{}.
\newblock \showarticletitle{Continuous-time dynamic network embeddings}. In
  \bibinfo{booktitle}{\emph{WWW}}. \bibinfo{pages}{969--976}.
\newblock


\bibitem[\protect\citeauthoryear{Perez, Strub, de~Vries, Dumoulin, and
  Courville}{Perez et~al\mbox{.}}{2018}]%
        {perez2018film}
\bibfield{author}{\bibinfo{person}{Ethan Perez}, \bibinfo{person}{Florian
  Strub}, \bibinfo{person}{Harm de Vries}, \bibinfo{person}{Vincent Dumoulin},
  {and} \bibinfo{person}{Aaron~C Courville}.} \bibinfo{year}{2018}\natexlab{}.
\newblock \showarticletitle{FiLM: Visual reasoning with a general conditioning
  layer}. In \bibinfo{booktitle}{\emph{AAAI}}. \bibinfo{pages}{3942--3951}.
\newblock


\bibitem[\protect\citeauthoryear{Perozzi, Al-Rfou, and Skiena}{Perozzi
  et~al\mbox{.}}{2014}]%
        {perozzi2014deepwalk}
\bibfield{author}{\bibinfo{person}{Bryan Perozzi}, \bibinfo{person}{Rami
  Al-Rfou}, {and} \bibinfo{person}{Steven Skiena}.}
  \bibinfo{year}{2014}\natexlab{}.
\newblock \showarticletitle{{DeepWalk}: Online learning of social
  representations}. In \bibinfo{booktitle}{\emph{KDD}}.
  \bibinfo{pages}{701--710}.
\newblock


\bibitem[\protect\citeauthoryear{Rosenstein, Marx, Kaelbling, and
  Dietterich}{Rosenstein et~al\mbox{.}}{2005}]%
        {rosenstein2005transfer}
\bibfield{author}{\bibinfo{person}{Michael~T Rosenstein},
  \bibinfo{person}{Zvika Marx}, \bibinfo{person}{Leslie~Pack Kaelbling}, {and}
  \bibinfo{person}{Thomas~G Dietterich}.} \bibinfo{year}{2005}\natexlab{}.
\newblock \showarticletitle{To transfer or not to transfer}. In
  \bibinfo{booktitle}{\emph{NeurIPS Workshop on Transfer Learning}}.
  \bibinfo{pages}{1--4}.
\newblock


\bibitem[\protect\citeauthoryear{Santoro, Bartunov, Botvinick, Wierstra, and
  Lillicrap}{Santoro et~al\mbox{.}}{2016}]%
        {santoro2016meta}
\bibfield{author}{\bibinfo{person}{Adam Santoro}, \bibinfo{person}{Sergey
  Bartunov}, \bibinfo{person}{Matthew Botvinick}, \bibinfo{person}{Daan
  Wierstra}, {and} \bibinfo{person}{Timothy Lillicrap}.}
  \bibinfo{year}{2016}\natexlab{}.
\newblock \showarticletitle{Meta-learning with memory-augmented neural
  networks}. In \bibinfo{booktitle}{\emph{ICML}}. \bibinfo{pages}{1842--1850}.
\newblock


\bibitem[\protect\citeauthoryear{Seah, Sun, and S~Bhowmick}{Seah
  et~al\mbox{.}}{2018}]%
        {seah2018killing}
\bibfield{author}{\bibinfo{person}{Boon-Siew Seah}, \bibinfo{person}{Aixin
  Sun}, {and} \bibinfo{person}{Sourav S~Bhowmick}.}
  \bibinfo{year}{2018}\natexlab{}.
\newblock \showarticletitle{Killing two birds with one stone: Concurrent
  ranking of tags and comments of social images}. In
  \bibinfo{booktitle}{\emph{SIGIR}}. \bibinfo{pages}{937--940}.
\newblock


\bibitem[\protect\citeauthoryear{Snell, Swersky, and Zemel}{Snell
  et~al\mbox{.}}{2017}]%
        {snell2017prototypical}
\bibfield{author}{\bibinfo{person}{Jake Snell}, \bibinfo{person}{Kevin
  Swersky}, {and} \bibinfo{person}{Richard Zemel}.}
  \bibinfo{year}{2017}\natexlab{}.
\newblock \showarticletitle{Prototypical networks for few-shot learning}. In
  \bibinfo{booktitle}{\emph{NIPS}}. \bibinfo{pages}{4077--4087}.
\newblock


\bibitem[\protect\citeauthoryear{Suo, Chou, Zhong, and Zhang}{Suo
  et~al\mbox{.}}{2020}]%
        {suo2020tadanet}
\bibfield{author}{\bibinfo{person}{Qiuling Suo}, \bibinfo{person}{Jingyuan
  Chou}, \bibinfo{person}{Weida Zhong}, {and} \bibinfo{person}{Aidong Zhang}.}
  \bibinfo{year}{2020}\natexlab{}.
\newblock \showarticletitle{TAdaNet: Task-adaptive network for graph-enriched
  meta-learning}. In \bibinfo{booktitle}{\emph{KDD}}.
  \bibinfo{pages}{1789--1799}.
\newblock


\bibitem[\protect\citeauthoryear{Sutherland, O'brien, and Weaver}{Sutherland
  et~al\mbox{.}}{2003}]%
        {sutherland2003spline}
\bibfield{author}{\bibinfo{person}{Jeffrey~J Sutherland},
  \bibinfo{person}{Lee~A O'brien}, {and} \bibinfo{person}{Donald~F Weaver}.}
  \bibinfo{year}{2003}\natexlab{}.
\newblock \showarticletitle{Spline-fitting with a genetic algorithm: A method
  for developing classification structure-activity relationships}.
\newblock \bibinfo{journal}{\emph{Journal of Chemical Information and Computer
  Sciences}} \bibinfo{volume}{43}, \bibinfo{number}{6} (\bibinfo{year}{2003}),
  \bibinfo{pages}{1906--1915}.
\newblock


\bibitem[\protect\citeauthoryear{Tang, Qu, Wang, Zhang, Yan, and Mei}{Tang
  et~al\mbox{.}}{2015}]%
        {tang2015line}
\bibfield{author}{\bibinfo{person}{Jian Tang}, \bibinfo{person}{Meng Qu},
  \bibinfo{person}{Mingzhe Wang}, \bibinfo{person}{Ming Zhang},
  \bibinfo{person}{Jun Yan}, {and} \bibinfo{person}{Qiaozhu Mei}.}
  \bibinfo{year}{2015}\natexlab{}.
\newblock \showarticletitle{LINE: Large-scale information network embedding}.
  In \bibinfo{booktitle}{\emph{WWW}}. \bibinfo{pages}{1067--1077}.
\newblock


\bibitem[\protect\citeauthoryear{Triantafillou, Zhu, Dumoulin, Lamblin, Evci,
  Xu, Goroshin, Gelada, Swersky, Manzagol, et~al\mbox{.}}{Triantafillou
  et~al\mbox{.}}{2019}]%
        {triantafillou2019meta}
\bibfield{author}{\bibinfo{person}{Eleni Triantafillou}, \bibinfo{person}{Tyler
  Zhu}, \bibinfo{person}{Vincent Dumoulin}, \bibinfo{person}{Pascal Lamblin},
  \bibinfo{person}{Utku Evci}, \bibinfo{person}{Kelvin Xu},
  \bibinfo{person}{Ross Goroshin}, \bibinfo{person}{Carles Gelada},
  \bibinfo{person}{Kevin Swersky}, \bibinfo{person}{Pierre-Antoine Manzagol},
  {et~al\mbox{.}}} \bibinfo{year}{2019}\natexlab{}.
\newblock \showarticletitle{Meta-Dataset: a dataset of datasets for learning to
  learn from few examples}. In \bibinfo{booktitle}{\emph{ICLR}}.
\newblock


\bibitem[\protect\citeauthoryear{Veli{\v{c}}kovi{\'c}, Cucurull, Casanova,
  Romero, Lio, and Bengio}{Veli{\v{c}}kovi{\'c} et~al\mbox{.}}{2018}]%
        {velivckovic2017graph}
\bibfield{author}{\bibinfo{person}{Petar Veli{\v{c}}kovi{\'c}},
  \bibinfo{person}{Guillem Cucurull}, \bibinfo{person}{Arantxa Casanova},
  \bibinfo{person}{Adriana Romero}, \bibinfo{person}{Pietro Lio}, {and}
  \bibinfo{person}{Yoshua Bengio}.} \bibinfo{year}{2018}\natexlab{}.
\newblock \showarticletitle{Graph attention networks}. In
  \bibinfo{booktitle}{\emph{ICLR}}.
\newblock


\bibitem[\protect\citeauthoryear{Velickovic, Fedus, Hamilton, Lio, Bengio, and
  Hjelm}{Velickovic et~al\mbox{.}}{2019}]%
        {velickovic2019deep}
\bibfield{author}{\bibinfo{person}{Petar Velickovic}, \bibinfo{person}{William
  Fedus}, \bibinfo{person}{William~L Hamilton}, \bibinfo{person}{Pietro Lio},
  \bibinfo{person}{Yoshua Bengio}, {and} \bibinfo{person}{R~Devon Hjelm}.}
  \bibinfo{year}{2019}\natexlab{}.
\newblock \showarticletitle{Deep graph infomax}. In
  \bibinfo{booktitle}{\emph{ICLR}}.
\newblock


\bibitem[\protect\citeauthoryear{Vinyals, Blundell, Lillicrap, Wierstra,
  et~al\mbox{.}}{Vinyals et~al\mbox{.}}{2016}]%
        {vinyals2016matching}
\bibfield{author}{\bibinfo{person}{Oriol Vinyals}, \bibinfo{person}{Charles
  Blundell}, \bibinfo{person}{Timothy Lillicrap}, \bibinfo{person}{Daan
  Wierstra}, {et~al\mbox{.}}} \bibinfo{year}{2016}\natexlab{}.
\newblock \showarticletitle{Matching networks for one shot learning}. In
  \bibinfo{booktitle}{\emph{NIPS}}. \bibinfo{pages}{3630--3638}.
\newblock


\bibitem[\protect\citeauthoryear{Wu, Souza~Jr, Zhang, Fifty, Yu, and
  Weinberger}{Wu et~al\mbox{.}}{2019}]%
        {wu2019simplifying}
\bibfield{author}{\bibinfo{person}{Felix Wu}, \bibinfo{person}{Amauri~H
  Souza~Jr}, \bibinfo{person}{Tianyi Zhang}, \bibinfo{person}{Christopher
  Fifty}, \bibinfo{person}{Tao Yu}, {and} \bibinfo{person}{Kilian~Q
  Weinberger}.} \bibinfo{year}{2019}\natexlab{}.
\newblock \showarticletitle{Simplifying graph convolutional networks}. In
  \bibinfo{booktitle}{\emph{ICML}}. \bibinfo{pages}{6861--6871}.
\newblock


\bibitem[\protect\citeauthoryear{Wu, Li, So, Wright, and Chang}{Wu
  et~al\mbox{.}}{2012}]%
        {wu2012learning}
\bibfield{author}{\bibinfo{person}{Xiao-Ming Wu}, \bibinfo{person}{Zhenguo Li},
  \bibinfo{person}{Anthony~M So}, \bibinfo{person}{John Wright}, {and}
  \bibinfo{person}{Shih-Fu Chang}.} \bibinfo{year}{2012}\natexlab{}.
\newblock \showarticletitle{Learning with partially absorbing random walks}. In
  \bibinfo{booktitle}{\emph{NIPS}}. \bibinfo{pages}{3086--3094}.
\newblock


\bibitem[\protect\citeauthoryear{Wu, Pan, Chen, Long, Zhang, and Philip}{Wu
  et~al\mbox{.}}{2020}]%
        {wu2020comprehensive}
\bibfield{author}{\bibinfo{person}{Zonghan Wu}, \bibinfo{person}{Shirui Pan},
  \bibinfo{person}{Fengwen Chen}, \bibinfo{person}{Guodong Long},
  \bibinfo{person}{Chengqi Zhang}, {and} \bibinfo{person}{S~Yu Philip}.}
  \bibinfo{year}{2020}\natexlab{}.
\newblock \showarticletitle{A comprehensive survey on graph neural networks}.
\newblock \bibinfo{journal}{\emph{IEEE TNNLS}}  \bibinfo{volume}{32}
  (\bibinfo{year}{2020}), \bibinfo{pages}{4--24}.
\newblock
Issue 1.


\bibitem[\protect\citeauthoryear{Xu, Ruan, Korpeoglu, Kumar, and Achan}{Xu
  et~al\mbox{.}}{2020}]%
        {Xu2020Inductive}
\bibfield{author}{\bibinfo{person}{Da Xu}, \bibinfo{person}{Chuanwei Ruan},
  \bibinfo{person}{Evren Korpeoglu}, \bibinfo{person}{Sushant Kumar}, {and}
  \bibinfo{person}{Kannan Achan}.} \bibinfo{year}{2020}\natexlab{}.
\newblock \showarticletitle{Inductive representation learning on temporal
  graphs}. In \bibinfo{booktitle}{\emph{ICLR}}.
\newblock


\bibitem[\protect\citeauthoryear{Xu, Hu, Leskovec, and Jegelka}{Xu
  et~al\mbox{.}}{2019}]%
        {xu2018powerful}
\bibfield{author}{\bibinfo{person}{Keyulu Xu}, \bibinfo{person}{Weihua Hu},
  \bibinfo{person}{Jure Leskovec}, {and} \bibinfo{person}{Stefanie Jegelka}.}
  \bibinfo{year}{2019}\natexlab{}.
\newblock \showarticletitle{How powerful are graph neural networks?}. In
  \bibinfo{booktitle}{\emph{ICLR}}.
\newblock


\bibitem[\protect\citeauthoryear{Yang, Cohen, and Salakhudinov}{Yang
  et~al\mbox{.}}{2016}]%
        {yang2016revisiting}
\bibfield{author}{\bibinfo{person}{Zhilin Yang}, \bibinfo{person}{William
  Cohen}, {and} \bibinfo{person}{Ruslan Salakhudinov}.}
  \bibinfo{year}{2016}\natexlab{}.
\newblock \showarticletitle{Revisiting semi-supervised learning with graph
  embeddings}. In \bibinfo{booktitle}{\emph{ICML}}. PMLR,
  \bibinfo{pages}{40--48}.
\newblock


\bibitem[\protect\citeauthoryear{Yao, Wei, Huang, and Li}{Yao
  et~al\mbox{.}}{2019}]%
        {yao2019hierarchically}
\bibfield{author}{\bibinfo{person}{Huaxiu Yao}, \bibinfo{person}{Ying Wei},
  \bibinfo{person}{Junzhou Huang}, {and} \bibinfo{person}{Zhenhui Li}.}
  \bibinfo{year}{2019}\natexlab{}.
\newblock \showarticletitle{Hierarchically Structured Meta-learning}. In
  \bibinfo{booktitle}{\emph{ICML}}. \bibinfo{pages}{7045--7054}.
\newblock


\bibitem[\protect\citeauthoryear{Yao, Zhang, Wei, Jiang, Wang, Huang, Chawla,
  and Li}{Yao et~al\mbox{.}}{2020}]%
        {yao2019graph}
\bibfield{author}{\bibinfo{person}{Huaxiu Yao}, \bibinfo{person}{Chuxu Zhang},
  \bibinfo{person}{Ying Wei}, \bibinfo{person}{Meng Jiang},
  \bibinfo{person}{Suhang Wang}, \bibinfo{person}{Junzhou Huang},
  \bibinfo{person}{Nitesh~V Chawla}, {and} \bibinfo{person}{Zhenhui Li}.}
  \bibinfo{year}{2020}\natexlab{}.
\newblock \showarticletitle{Graph few-shot learning via knowledge transfer}. In
  \bibinfo{booktitle}{\emph{AAAI}}. \bibinfo{pages}{6656--6663}.
\newblock


\bibitem[\protect\citeauthoryear{You, Ying, and Leskovec}{You
  et~al\mbox{.}}{2019}]%
        {you2019position}
\bibfield{author}{\bibinfo{person}{Jiaxuan You}, \bibinfo{person}{Rex Ying},
  {and} \bibinfo{person}{Jure Leskovec}.} \bibinfo{year}{2019}\natexlab{}.
\newblock \showarticletitle{Position-aware graph neural networks}. In
  \bibinfo{booktitle}{\emph{ICML}}. \bibinfo{pages}{7134--7143}.
\newblock


\bibitem[\protect\citeauthoryear{Zeng, Zhou, Srivastava, Kannan, and
  Prasanna}{Zeng et~al\mbox{.}}{2020}]%
        {Zeng2020GraphSAINT:}
\bibfield{author}{\bibinfo{person}{Hanqing Zeng}, \bibinfo{person}{Hongkuan
  Zhou}, \bibinfo{person}{Ajitesh Srivastava}, \bibinfo{person}{Rajgopal
  Kannan}, {and} \bibinfo{person}{Viktor Prasanna}.}
  \bibinfo{year}{2020}\natexlab{}.
\newblock \showarticletitle{GraphSAINT: Graph sampling based inductive learning
  method}. In \bibinfo{booktitle}{\emph{ICLR}}.
\newblock


\bibitem[\protect\citeauthoryear{Zhang, Liu, Tang, Chen, and Li}{Zhang
  et~al\mbox{.}}{2013}]%
        {zhang2013social}
\bibfield{author}{\bibinfo{person}{Jing Zhang}, \bibinfo{person}{Biao Liu},
  \bibinfo{person}{Jie Tang}, \bibinfo{person}{Ting Chen}, {and}
  \bibinfo{person}{Juanzi Li}.} \bibinfo{year}{2013}\natexlab{}.
\newblock \showarticletitle{Social influence locality for modeling retweeting
  behaviors}. In \bibinfo{booktitle}{\emph{IJCAI}}, Vol.~\bibinfo{volume}{13}.
  \bibinfo{pages}{2761--2767}.
\newblock


\bibitem[\protect\citeauthoryear{Zhou, Bousquet, Lal, Weston, and
  Sch{\"o}lkopf}{Zhou et~al\mbox{.}}{2003}]%
        {zhou_learning_2003}
\bibfield{author}{\bibinfo{person}{Dengyong Zhou}, \bibinfo{person}{Olivier
  Bousquet}, \bibinfo{person}{Thomas~Navin Lal}, \bibinfo{person}{Jason
  Weston}, {and} \bibinfo{person}{Bernhard Sch{\"o}lkopf}.}
  \bibinfo{year}{2003}\natexlab{}.
\newblock \showarticletitle{Learning with local and global consistency}. In
  \bibinfo{booktitle}{\emph{{NIPS}}}. \bibinfo{pages}{321--328}.
\newblock


\bibitem[\protect\citeauthoryear{Zhou, Cao, Zhang, Trajcevski, Zhong, and
  Geng}{Zhou et~al\mbox{.}}{2019}]%
        {zhou2019meta}
\bibfield{author}{\bibinfo{person}{Fan Zhou}, \bibinfo{person}{Chengtai Cao},
  \bibinfo{person}{Kunpeng Zhang}, \bibinfo{person}{Goce Trajcevski},
  \bibinfo{person}{Ting Zhong}, {and} \bibinfo{person}{Ji Geng}.}
  \bibinfo{year}{2019}\natexlab{}.
\newblock \showarticletitle{Meta-GNN: On few-shot node classification in graph
  meta-learning}. In \bibinfo{booktitle}{\emph{CIKM}}.
  \bibinfo{pages}{2357--2360}.
\newblock


\bibitem[\protect\citeauthoryear{Zhu, Ghahramani, and Lafferty}{Zhu
  et~al\mbox{.}}{2003}]%
        {zhu_semi-supervised_2003}
\bibfield{author}{\bibinfo{person}{Xiaojin Zhu}, \bibinfo{person}{Zoubin
  Ghahramani}, {and} \bibinfo{person}{John Lafferty}.}
  \bibinfo{year}{2003}\natexlab{}.
\newblock \showarticletitle{Semi-supervised learning using {Gaussian} fields
  and harmonic functions}. In \bibinfo{booktitle}{\emph{{ICML}}}.
  \bibinfo{pages}{912--919}.
\newblock


\bibitem[\protect\citeauthoryear{Zuo, Liu, Lin, Guo, Hu, and Wu}{Zuo
  et~al\mbox{.}}{2018}]%
        {zuo2018embedding}
\bibfield{author}{\bibinfo{person}{Yuan Zuo}, \bibinfo{person}{Guannan Liu},
  \bibinfo{person}{Hao Lin}, \bibinfo{person}{Jia Guo},
  \bibinfo{person}{Xiaoqian Hu}, {and} \bibinfo{person}{Junjie Wu}.}
  \bibinfo{year}{2018}\natexlab{}.
\newblock \showarticletitle{Embedding temporal network via neighborhood
  formation}. In \bibinfo{booktitle}{\emph{KDD}}. \bibinfo{pages}{2857--2866}.
\newblock


\end{thebibliography}


\end{document}